\documentclass[12pt]{article}
\usepackage{coling2018}
\usepackage{times}
\usepackage[utf8]{inputenc}
\usepackage{setspace}
\usepackage{url}
\usepackage{latexsym}
\usepackage{graphicx}
\usepackage{adjustbox}
\usepackage{amsfonts}
\usepackage{booktabs}
\usepackage{subcaption}
\usepackage{wrapfig}
\usepackage{phaistos}
\usepackage{float}
\usepackage{tipa}
\usepackage[margin=1in]{geometry}

\usepackage{multirow}
\usepackage{array}
\newcolumntype{P}[1]{>{\centering\arraybackslash}p{#1}}
\usepackage{longtable}

\usepackage[linkbordercolor={0 0 1}]{hyperref}
\hypersetup{
    colorlinks=true,
    linkcolor=blue,
    filecolor=blue,      
    urlcolor=blue,
    citecolor=blue,
    pdftitle={Sharelatex Example},
    pdfpagemode=FullScreen,
}

\usepackage{makecell}
\usepackage{lingmacros}
\usepackage{cellspace, tabularx, array}
\usepackage[disable]{todonotes}
\newcommand{\note}[4][]{\todo[author=#2,color=#3,size=\scriptsize,fancyline,caption={},#1]{#4}}

\newcommand{\mans}[2][]{\note[#1]{MH}{blue!40}{#2}{}}

\usepackage{fancyhdr}
\renewcommand{\headrulewidth}{0pt} 
\title{Computational Morphology with Neural Network Approaches}

\author{Ling Liu \\ University of Colorado \\ ling.liu@colorado.edu}

\begin{document}
\maketitle

\begin{abstract}
Neural network approaches have been applied to computational morphology with great success, improving the performance of most tasks by a large margin and providing new perspectives for modeling. This paper starts with a brief introduction to computational morphology, followed by a review of recent work on computational morphology with neural network approaches, to provide an overview of the area. In the end, we will analyze the advantages and problems of neural network approaches to computational morphology, and point out some directions to be explored by future research and study.
\end{abstract}

\section{Introduction}
\label{sec:intro}

In the word $drivers$, there are three parts which can't be split further without resulting in meaningless units in English: $drive$, \textit{-}$(e)r$, and \textit{-}$s$.\footnote{``\textit{-}'' is used to indicate that the string is not a word by itself, \textit{-string} indicates the string is usually added to the end of a word, i.e. it's a suffix; \textit{string-} indicates the string is usually added to the beginning of a word, i.e. it's a prefix; and \textit{-string-} indicates the string is usually inserted into a word, i.e. it's an infix.} The first part $drive$ is a verb, meaning to operate to make vehicles move; the second part \textit{-}$(e)r$ has the function of changing the verb to a noun meaning the person who does the actions; and the third part \textit{-}$s$ has the function of indicating there are more than one of this noun. Analyzing the internal structure of words, understanding the meaning and function related to each part, and figuring out how different units can be combined to make valid words, is the focus of the area of linguistic study known as morphology. According to \newcite{haspelmath2013understanding}, ``Morphology is the study of systematic covariation in the form and meaning of words.'' (p2) Note that this definition and the example in the beginning presupposes that language can be divided into distinct units of word. This paper focuses on studies which follow this presupposition and use data clearly marked with word boundaries. In other words, we focus on data where larger units have been segmented into words either with natural word boundaries like spaces in English or with some preprocessing for languages without natural word boundaries like Chinese and Japanese. 

In the example of $drivers$, the three parts are three different morphemes, the first morpheme $drive$ is a word in English which can't the divided further, also called a stem, the other two parts which are not words but can be attached to stems or larger units are named affixes. There are two types of processes involved to form the word \textit{drivers}: derivation and inflection. The process of concatenating $drive$ and \textit{-}$(e)r$ to form $driver$ is a derivational process, where the new word has a different concrete lexical meaning; the process of adding \textit{-}$s$ to the end of $driver$ to get the word form \textit{drivers} is more related to the grammatical requirements by the English language, and this process is known as inflection.

Computational methods can be applied to facilitate morphological study, which is the focus of computational morphology. For example, finite state machines and two-level morphology provide ways to formulate the morphological rules linguists come up with to generalize over word formation in different languages and to use these rules to automatically analyze or generate more data. Finite state machines and two-level morphology are very linguistics-oriented and rely directly on linguistic studies. Machine learning techniques like Hidden Markov Models (HMM), Conditional Random Fields (CRFs), Support Vector Machines (SVMs) do not rely on linguistics as much. These machine learning techniques provide ways for the algorithm to learn about morphology from annotated data or raw text without explicit linguistic rules, though they still require high-level feature engineering and usually use heuristics designed with heavy linguistic considerations. Neural network models have been flourishing in recent years and generated state-of-the-art results for many tasks in natural language processing (NLP), including computational morphology. Since \newcite{kann2016med} applied the neural encoder-decoder models to SIGMORPHON 2016 shared task on morphological (re)inflection \cite{cotterell2016sigmorphon}, neural network models have been dominant in computational morphology and achieved significant improvements over previous results. Such models rely heavily on large amounts of annotated data in order to achieve good performance, but they don't require explicit linguistic rules or high-level feature engineering. Being free from feature engineering is an advantage from the engineering side, but is a disadvantage from the linguistics side because it's hard to interpret what the models learn, making it less helpful to contribute to the understanding of human languages. 

From the NLP perspective, computational morphology has been an important data preprocessing step for downstream tasks like machine translation, information retrieval, dependency parsing etc. Though various ways to make use of subword information like characters, n-grams, or byte pair encodings have been explored, high-quality morphological processing remains most helpful, especially for morphologically rich languages (MRLs)\footnote{\newcite{conforti-etal-2018-neural} define MRL as ``a language where word shapes encode a consistent number of syntactic and semantic features'', like fusional languages such as Czech, and agglutinating languages such as Finnish, Hungarian, and Turkish.}  \cite{belinkov-etal-2017-neural,vania-etal-2018-character,dehouck-denis-2018-framework,klein-tsarfaty-2020-getting}. 

The overwhelmingly good performance of neural network models indicates that they are good models for NLP purposes, and they can't be ignored either for linguistic and cognitive understanding of language. With such a consideration, this paper will review recent work (mainly work published between 2016 and 2020) using \textbf{language-agnostic} neural network models\footnote{Language-agnostic models refer to models which have been developed for and tested on multiple languages without being engineered for each language specifically.} to process computational morphology \textbf{with the focus on inflection}. We will summarize what tasks have been using such models, what architectures the neural models have and what results have been generated. Then we will analyze the advantages and disadvantages of the neural models for the tasks, and point out some directions for future work. 

The remaining of the paper is organized as follows: Section \ref{sec:tasks} will define tasks in computational morphology. Section \ref{sec:neural_models} will introduce basic neural network architectures, which have been combined with each other or other machine learning approaches to process computational morphology. Sections \ref{sec:type-based} and \ref{sec:token-based} will explain in more detail how different models have been applied to different tasks. 
Section \ref{sec:discuss} will discuss the advantages and problems of neural network approaches, and point out some directions for future research and study. This paper will end with a conclusion in Section \ref{sec:conclude}.

\section{Tasks in computational morphology}
\label{sec:tasks}

\begin{figure}
\centering{\includegraphics[width=0.8\textwidth]{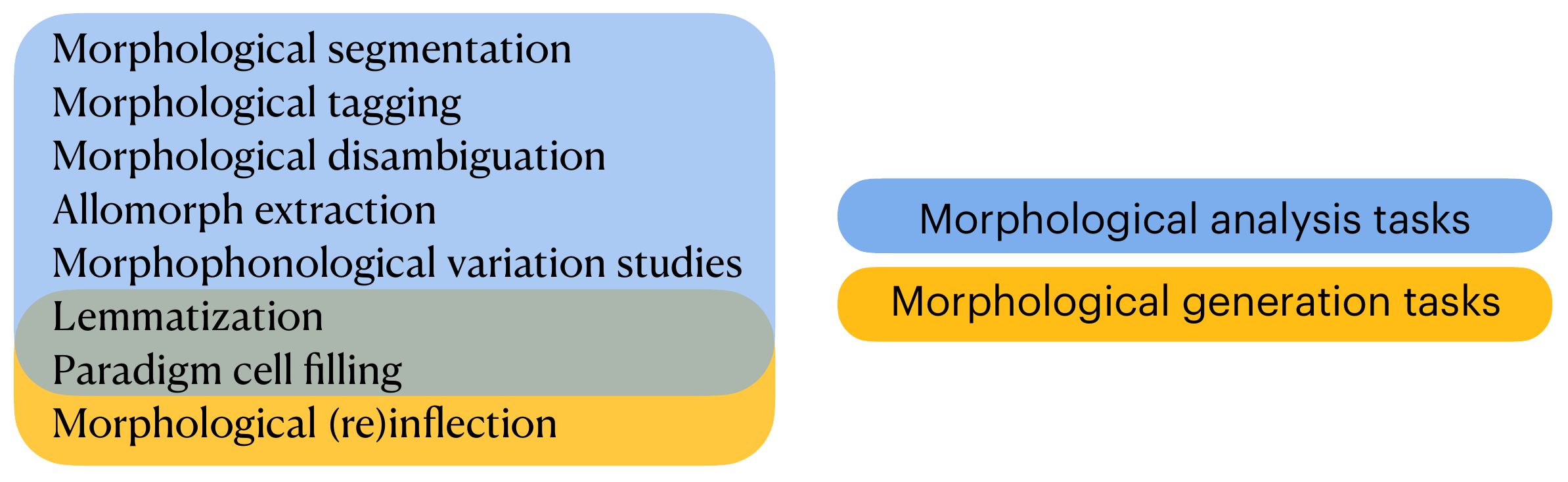}}
\caption{Summary of morphological learning tasks. Analysis tasks are shaded in blue (the top box), and generation tasks are shaded in orange (the bottom box). Lemmatization and paradigm cell filling (the overlap part between the top and bottom boxes) involve both analysis and generation.}
\label{fig:morph_tasks}
\end{figure}

Tasks in computational morphology (summarized in Figure \ref{fig:morph_tasks}) can be generally classified into two types: analysis and generation. For morphological analysis, the goal is to learn about the morphological structure of a given word form. Analysis tasks include morphological tagging, morphological segmentation, morphological disambiguation, allomorphy extraction and morphophonological variation studies. For morphological generation, the goal is to generate a correct word form, including tasks like inflection, reinflection etc. When the reinflection is to generate the lemma form (i.e. the base or dictionary form) of a word given its other inflected forms, the task is called lemmatization. However, lemmatization is also closely related to morphological analysis: morphological tagging and morphological segmentation are usually indispensable from lemmatization. Paradigm completion tasks may also involve both analysis and generation processes. When the task is conducted on the lexicon, i.e. word in isolation, we call it a \textbf{type-based} task; and when the task is about words in context, usually sentential context, we call it a \textbf{token-based} task.

\subsection{Morphological analysis}

The term \textit{morphological analysis} has been used quite generally to refer to all the morphological learning tasks to analyze or understand the given word form. This term has also been used in a more specific sense, where it refers to the task of ``[annotating] a given word form with its lemma and morphological tag'' \cite{nicolai-kondrak-2017-morphological} (p211), i.e. the combination of lemmatization and morphological tagging. In this paper, we use \textit{morphological analysis} to refer to these two broader and narrower senses. In the literature, \textit{morphological analysis} has also been used interchangeably with \textit{morphological tagging} to refer to the task of ``predicting fine-grained annotations about the syntactic properties of tokens in a language such as part-of-speech, case, or tense'' \cite{malaviya-etal-2018-neural} (p2653), but this paper avoids this usage. 

\subsubsection{Morphological tagging and lemmatization}

\begin{figure}
\centering{\includegraphics[width=.8\textwidth]{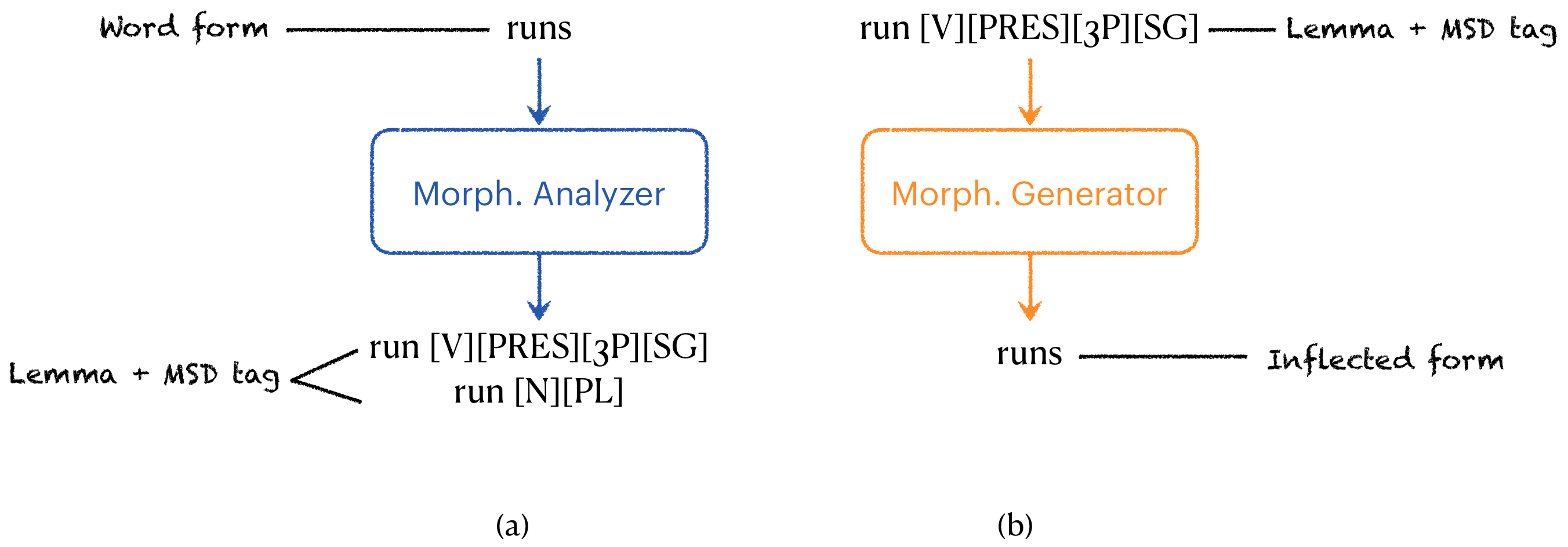}}
\caption{(a) is a morphological analyzer which conducts lemmatization and morphological tagging on the English word \textit{runs}. (b) is a morphological generator which produces the inflected form corresponding to the English lemma \textit{run} and the MSD tag \textit{V;PRES;3P;SG}.}
\label{fig:morph_analyzer_generator}
\end{figure}

\textit{Morphological tagging} is closely related to part-of-speech (POS) tagging, and has been treated as fine-grained POS tagging \cite{labeau-etal-2015-non,conforti-etal-2018-neural} or a generalization of POS tagging \cite{cotterell-heigold-2017-cross}. A tag for morphological description usually consists of a POS label together with more specific labels which describe the morphosyntactic features of each word. Such morphological tags have been referred to as MSDs (morphosyntactic descriptions), features, labels, or tags. In this remainder of the paper, we will use \textbf{MSD tag} to refer to the whole chunk of labels corresponding to each word, e.g. V;PRES;3P;SG for \textit{runs}, and refer to each component label like V, PRES, 3P, SG as its \textbf{features}. \textit{Lemmatization} is the task to convert the word to the normalized form (i.e. lemma form), which is usually the base or dictionary form of the word \cite{plisson2004rule,bergmanis2018context}. Figure \ref{fig:morph_analyzer_generator}(a)\footnote{The figure is cited from Hulden's presentation slides (2015).} illustrates a typical type-based morphological analyzer which conducts \textit{lemmatization} and \textit{morphological tagging} where the input is a word form and the morphological analyzer is expected to produce all the possible lemmata with the MSD tags corresponding to the word form. 
\textbf{Context} can \textit{disambiguate} lemmatization and tagging as well as improve accuracy on ambiguous and unseen words \cite{bergmanis2018context,bergmanis-goldwater-2019-training} because semantic meaning and morphosyntactic features can be inferred from contextual information to some extent. The second subtask in CoNLL-SIGMORPHON 2019 shared task \cite{mccarthy-etal-2019-sigmorphon} is about this: \textit{morphological tagging} and \textit{lemmatization} in context. Specifically, the input is a sentence, and the output is expected to be the corresponding lemma and MSD tag for each word in the sentence. See Table \ref{tab:contextual_lemmatization_tagging_example} for an example.

\bigbreak

\bigbreak

\begin{table}
    \centering
    \begin{adjustbox}{width=\textwidth}
    \begin{tabular}{llcccccc}
        \toprule
        INPUT & Sentence & He & runs & a & investment & company & . \\
        OUTPUT& Lemmatization & he & run & a & investment & company & . \\
        OUTPUT& Morph. Tagging & PRON;NOM;SG & V;SG;3;PRS & DET;IND & N;SG & N;SG & PUNCT \\
        \bottomrule
    \end{tabular}
    \end{adjustbox}
    \caption{Example of lemmatization and morphological tagging in context.}
    \label{tab:contextual_lemmatization_tagging_example}
\end{table}

\subsubsection{Morphological segmentation}

\textit{Morphological segmentation} is ``the problem of how to split each word up into appropriate functional subparts'' \cite{goldsmith2017computational} (p91). Not every word is separable without fuzziness. There are four cases of separability as illustrated with the following examples where $+$ is used to indicate morpheme boundaries: 
\begin{enumerate}
    \item Completely separable, e.g. $played \rightarrow play + ed$ \footnote{When the format $X \rightarrow Y$ is used, it means the input to the task is $X$, the (actual or expected) output is $Y$, and $\rightarrow$ can be thought of as the model which takes the input and produces the output.}.
    \item Separable, but with word form changes, though no change in the stem meaning, e.g. $sitting$ can be segmented in three different ways: (1) $sitting \rightarrow sit + ting$ (2) $sitt + ing$ (3) $sit + t + ing$.
    \item Separable, but there are multiple ways of segmentation, and different segmentations result in different stems and meanings, e.g. the Turkish word \textit{dolar} can be separated in four different valid ways as shown in Table \ref{tab:TurkishSegmentation}; 
    \item Not really separable, e.g. it's questionable to segment $swam$ like this: $swam \rightarrow swim + ed$, and we don't have a better alternative.
\end{enumerate}

\begin{table}
    \centering
    \begin{adjustbox}{width=.9\textwidth}
    \begin{tabular}{llll}
    \toprule
        Segmentation & Tagging & Lemmatization & English Translation \\
        \midrule
        dolar & N;3SG;Pnon;Nominative & dolar & dollar \\
        dola+r & V;Positive;Aorist;3sg & dola & he/she wraps \\
        dol+ar & V;Positive;Aorist;3sg & dol & it fills \\
        do+lar & N;3PL;Pnon;Nominative & do & Multiple C (musical note) \\
    \bottomrule
    \end{tabular}
    \end{adjustbox}
    \caption{Four different ways to segment of the Turkish word \textit{dolar}. \textit{``+''} marks morpheme boundaries. Each of \textit{dolar}, \textit{dola}, \textit{dol}, and \textit{do} is a valid root in Turkish and $\emptyset$, \textit{-r}, \textit{-ar}, \textit{-lar} are valid suffixes. $\emptyset$ means empty string. The example is adopted from \protect\newcite{yildiz2016disambiguation}.}
    \label{tab:TurkishSegmentation}
\end{table}

The ambiguity and separability of the word also depends on the language. \newcite{ruokolainen-etal-2016-comparative} point out that ``[m]orphological segmentation can be most naturally applied to highly agglutinative languages'' (p95) in which multiple morphemes are usually concatenated together with clear boundaries.

For the second case of separability, the segmentation to keep the surface variations in the morphemes is \textbf{surface segmentation}, and if the changes to the morphemes during word formation are restored, it is \textbf{canonical segmentation} \cite{cotterell2016joint}.

\begin{figure}
\centering{\includegraphics[width=0.5\textwidth]{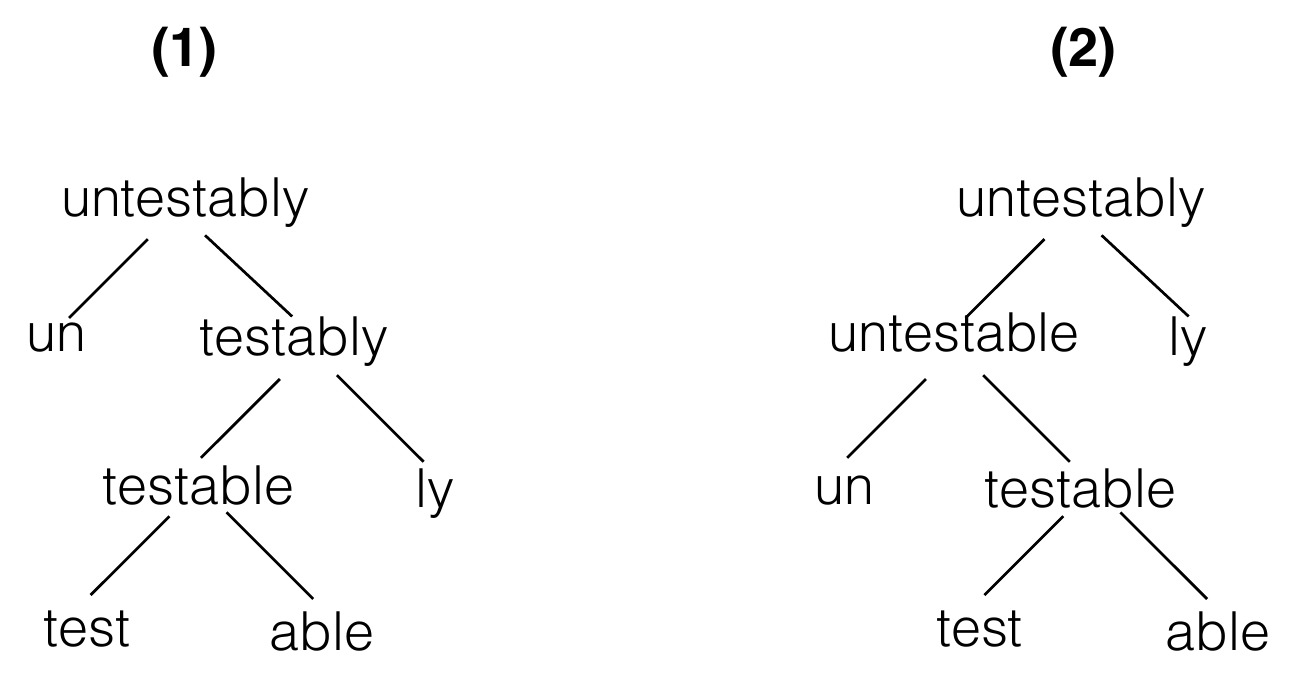}}
\caption{Parsing trees of two hierarchical segmentations of $untestably$. The trees were created by reference to \protect\newcite{cotterell-etal-2016-morphological-segmentation}}\label{fig:hierarchigalSeg}
\end{figure}

There are two different segmentation mechanisms: flat segmentation and hierarchical segmentation. This distinction is more related to the derivational than inflectional process. For example, for the word $untestably$, there is only one way of \textbf{flat segmentation}: $un + test + able + ly$, but there are two different ways of \textbf{hierarchical segmentation}: $[un[test[able[ly]]]]$ and $[[un[test[able]]]ly]$, which are two different parsing tree structures as illustrated in Figure \ref{fig:hierarchigalSeg} \cite{cotterell-etal-2016-morphological-segmentation}. The semantic ambiguity in the flat segmentation of the word can be clarified in the hierarchical segmentation \cite{cotterell-etal-2016-morphological-segmentation,steiner-2019-augmenting}.

\subsubsection{Morphological disambiguation}

In cases where multiple segmentations or analyses with different semantic and functional units are equally justified if the word is treated in isolation, like the Turkish example of \textit{dolar} in Table \ref{tab:TurkishSegmentation} and the English example of \textit{runs} in Figure \ref{fig:morph_analyzer_generator}(a), alternative segmentations or analyses are usually provided. If the context information is available, it can be applied to select the segmentation or lemma and MSD tag that work best for the context. This involves the task of \textit{morphological disambiguation}: to pick out the segmentation or lemma and tag of a given word form that matches the context. For example, the following two example sentences containing \textit{runs} are selected from \textit{COCA}.\footnote{\url{https://www.english-corpora.org/coca/}} A morphological analyzer as the one in Figure \ref{fig:morph_analyzer_generator}(a) returns two possible analyses for the word form \textit{runs}: (a) \textit{run [V][PRES][3P][SG]}, (b) \textit{run [N][PL]}, and a morphological disambiguator is expected to figure out that (a) matches the second sentence and it is (b) that is used in the first sentence.

\begin{singlespace}
\enumsentence{\textit{Seeing those two home \textbf{runs} makes you question anybody who says baseball can't be exciting.}}
\enumsentence{\textit{He \textbf{runs} an investment company.}}
\end{singlespace}

\newcite{cotterell-etal-2018-unsupervised} define a novel morphological disambiguation task, which they refer to as \textit{disambiguation of syncretism in inflected lexicons}. Syncretism is the phenomenon where the same word form falls into multiple slots within one paradigm. For example, \textit{run} is the non-3\textit{rd}-person-singular form of the lemma \textit{run} in present tense and the past participle form as well as the infinitive form of the lemma. 
Syncretism has been one of the cases neural models can't handle very well. What \newcite{cotterell-etal-2018-unsupervised} did with syncretism disambiguation is to partition the count of different morphological analyses for the same given word form in the corpus. For example, suppose the word form \textit{runs} appears in the corpus $X$ times in total, $Y$ out of the $X$ times it is used as \textit{run N;PL} and $Z$ out of the $X$ times it is used as \textit{run V;PRES;3P;SG}, the task of synchratism disambiguation is to estimate the fractional count of the two different analyses of the word form \textit{runs}, i.e. $Y/X$ for \textit{run N;PL} and $Z/X$ for \textit{run V;PRES;3P;SG}.

\subsubsection{Learning of allomorphy and morphophonological variations}
\label{subsubsec:allomorph}

There are two tasks that have been intertwined with morphological segmentation at the phonology-morphology interface, i.e. \textit{the learning of allomorphy and morphophonological variations}. The goal of these two task is to figure out how the various surface forms of morphological segments are related, including what are the underlying forms and what phonological changes are involved to derive the surface form from the underlying form \cite{cotterell2015modeling}. Table \ref{tab:EnglishSegmentation} provides the example of the regular plural suffixes for English nouns, which is realized as two orthographic allomorphs (i.e. \textit{-s} and \textit{-es} as shown in the column of Word segmentation) and three phonemic allomorphs (i.e. \textit{-s}, \textit{-z} and \textit{-\textipa{1z}} as shown in the column of IPA segmentation). The task of figuring out what morphological segments belong to the same group, and often also picking out an underlying form for them, is \textit{allomorphy extraction}. 

The task of \textit{morphophonological variation studies} is to handle and explain the variations in the surface forms. The example of English regular plural nouns involves voicing assimilation in the case of \textit{cats} and \textit{dogs}, and insertion of \textit{\textipa{1}} in the case of \textit{fishes}. There are much fewer publications for this task, perhaps because computational processing of morphology usually deals with orthographic representation of words rather than phonemic representations.

\begin{table}
    \centering
    \begin{adjustbox}{width=.78\textwidth}
    \begin{tabular}{lcccc}
    \toprule
       Word &  IPA transcription & Lemma & Word segmentation & IPA segmentation \\
    \midrule
       cats & \textipa{k\ae ts} & cat & cat+s & \textipa{k\ae t}+s \\
       dogs & \textipa{dAgz} & dog & dog+s & \textipa{dAg}+z \\
       fishes & \textipa{fIS1z} & fish & fish+es & \textipa{fIS}+\textipa{1z} \\
    \bottomrule
    \end{tabular}
    \end{adjustbox}
    \caption{Segmentation of English plural nouns \textit{cats}, \textit{dogs} and \textit{fishes}}
    \label{tab:EnglishSegmentation}
\end{table}

\subsection{Morphological generation}

\textit{Morphological generation} is a kind of string transduction or sequence-to-sequence transduction \cite{nicolai2015inflection,rastogi2016weighting,nicolai-etal-2018-string,ribeiro-etal-2018-local,makarov2018neural,makarov2018imitation,wu-etal-2018-hard}. When labels are provided, the task is called \textit{labeled sequence transduction} \cite{zhou2017multi}.

\begin{table}
    \centering
    \begin{adjustbox}{width=.78\textwidth}
    \begin{tabular}{l|ccc|c}
    \toprule
       & \multicolumn{3}{c|}{\textbf{INPUT}} & \textbf{OUTPUT} \\
       \textbf{Task}  &  \textbf{source form} & \textbf{source tag} & \textbf{target tag} & \textbf{target form} \\
    \midrule
        \textbf{Inflection} & run & - & V;3;SG;PRS & runs \\
        \textbf{Reinflection} & running & (V;V.PTCP;PRS) & V;3;SG;PRS & runs \\
        \textbf{Lemmatization} & running & (V;V.PTCP;PRS) & (V;NFIN) & run \\
    \bottomrule
    \end{tabular}
    \end{adjustbox}
    \caption{Examples of morphological inflection, reinflection, and lemmatization as a special case of reinflection. Tags in parenthesis may or may not be provided, and when they are provided, it is ``labeled sequence transduction'' \protect\cite{zhou2017multi}. ``-'' indicates the information is usually not specified.}
    \label{tab:generation_examples}
\end{table}

\subsubsection{Morphological (re)inflection}

Morphological generation is the inverse of morphological analysis, as illustrated in Figure \ref{fig:morph_analyzer_generator} on page \pageref{fig:morph_analyzer_generator}. In Figure \ref{fig:morph_analyzer_generator}(b), the lemma form of the English verb \textit{run} together with the MSD tag to be generated (\texttt{target tag}) is specified. The morphological generator is designed to produce the corresponding inflected form (\texttt{target form}). This task is \textit{morphological inflection} (i.e. \texttt{lemma + target tag $\rightarrow$ target form}). 
If the given word form (\texttt{source form}) is not limited to the lemma, the task becomes \textit{morphological reinflection} \cite{cotterell-etal-2017-conll}. For the morphological reinflection task, the MSD tag of the given word form (\texttt{source tag}) may or may not be provided (i.e. \texttt{source form + source tag + target tag $\rightarrow$ target form}, or \texttt{source form + target tag $\rightarrow$ target form}). 
\textit{Lemmatization} can be considered as a special case of \textit{morphological reinflection}, where the form to be generated is limited to the lemma form (i.e. \texttt{source form + source tag $\rightarrow$ lemma}, or \texttt{source form $\rightarrow$ lemma}). Table \ref{tab:generation_examples} provides examples for the three tasks. 



All the tasks can be \textbf{type-based}, i.e. to operate on the lexicon. In such tasks people have been working on, the MSD tags are usually explicitly specified. Annotated data at a relatively large scale is usually critical.  
In addition, semantic meaning and morphosyntactic features can be inferred from contextual information to some extent. \textbf{Inferring semantic and morphosyntactic information from context} makes it possible to train morphological inflection and reinflection models without relying on data explicitly annotated with MSD tags. For example, the second track of subtask 2 in CoNLL-SGIMORPHON 2018 shared task \cite{cotterell2018conll} is to generate an inflected word form given the lemma of the word and the context it occurs in, as exemplified with sentence (3) where the expected form is \textit{dogs} and the model is expected to generate this plural form of the noun:

\enumsentence{\textit{The $\rule{1cm}{0.15mm}$ (dog) are barking.}}

\begin{figure}
    \centering
    \includegraphics[width=.85\textwidth]{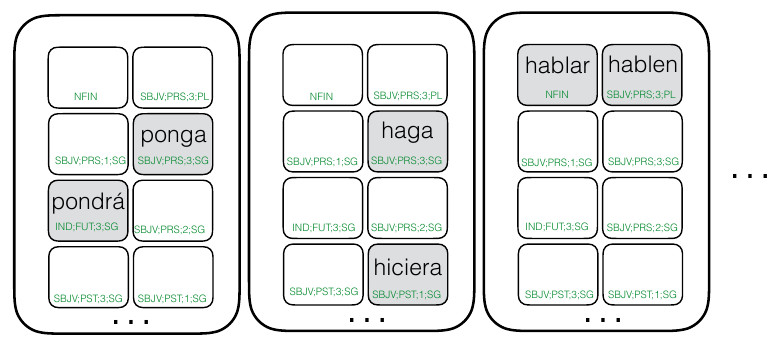}
    \caption{Illustration of the partial paradigm completion task. The examples are a fraction of Spanish verb paradigms. The PCFP task is to fill in all the missing slots in the given partially filled paradigms \protect\cite{silfverberg-hulden-2018-encoder}.}
    \label{fig:pcfp}
\end{figure}

\subsubsection{Paradigm completion}
\label{subsubsec:pcfp}

Another morphological generation task is \textit{paradigm completion}, also called \textit{the partial paradigm cell filling problem} (PCFP) \cite{ackerman2009}. 
In this task, incomplete inflection paradigms are provided, and the task is to fill in the missing cells in the partially filled inflection tables. Figure \ref{fig:pcfp} illustrates the task where incomplete Spanish inflection tables are provided and the task is to fill in all the missing slots in all the paradigms. 
Such a task can be reduced to the (re)inflection task by using the given slots to predict each missing slot when there are more than one slot given in the paradigm. However, if each partial paradigm has only one given slot, the problem is more much challenging and requires a different approach.  



\subsection{Other tasks}

In addition to the morphological analysis and generation tasks introduced above centering mainly on inflection, there are tasks involving \textbf{derivational morphology} as well \cite{cotterell-etal-2017-paradigm,vylomova-etal-2017-context,deutsch-etal-2018-distributional,hofmann-etal-2020-graph,hofmann-etal-2020-dagobert}, though derivation is much less studied than inflection in computational morphology. 

\textit{Historical text normalization} \cite{bollmann2016improving,bollmann2017learning,korchagina2017normalizing,robertson2018evaluating} aims to ``convert historical word forms to their modern equivalents'' \cite{robertson2018evaluating} (p720). This is also a task involving morphology. For example, the English word \textit{said} may be spelled as \textit{sayed}, \textit{seyd}, \textit{said}, \textit{sayd}, etc., and the historical text normalization system is expected to normalize these different spellings to the modern spelling \textit{said}, so that the historical text is more searchable and easier to process for downstream NLP tasks.

Deep learning models also provide new approaches to \textbf{simulate cognitive processing of morphology} and \textbf{quantify linguistic phenomena}. \newcite{kirov-cotterell-2018-recurrent} compare the learning curve of recurrent neural network models with cognitive linguistic findings of human language acquisition. \newcite{cotterell2018diachronic} investigate and model under what conditions irregular inflections can survive in a language over the course of time. \newcite{williams-etal-2019-quantifying} and \newcite{mccarthy-etal-2020-measuring} attempt to quantify the grammatical gender system across languages.

\subsection{Computational morphology: supervised, semi-supervised, unsupervised or reinforcement learning}

Another way to categorize tasks in machine learning, including computational morphology tasks, is by the use of annotated data. If a task relies on annotated data for training, it is a supervised task. If a task needs some annotated data, but can also use raw data or data not directly annotated for the task, it is semi-supervised. Unsupervised tasks operate on unannotated data like text crawled from Wikipedia. Reinforcement learning learns by taking actions in an environment to maximize cumulative reward. Currently there are few applications of reinforcement learning techniques to computational morphology.

\section{Neural approaches for morphology processing}
\label{sec:neural_models}

\subsection{Three learning scenarios}



The general setup of neural network models for computational morphology, is first to represent individual characters or other symbols depending on the desired level of granularity as one-hot vectors which are the inputs to the model, and then pass the one-hot vectors to an embedding layer. At the embedding layer, the model learns a weight matrix to convert each one-hot vector representation to a dense vector representation. 
The conversion from the input vector to the output vector is either through several steps or more simply in a single step. It's more common to have multiple steps and layers, and thus neural networks are usually called deep neural networks (DNNs). Based on how many output symbols are expected in relation to the number of input symbols, all the learning tasks can be generally interpreted into three different scenarios: many-to-one, many-to-many, and one-to-one scenarios \cite{bjerva2017one}. Different neural network architectures can be combined in different ways in each of the scenarios.

\begin{figure}
    \centering
    \includegraphics[width=.9\textwidth]{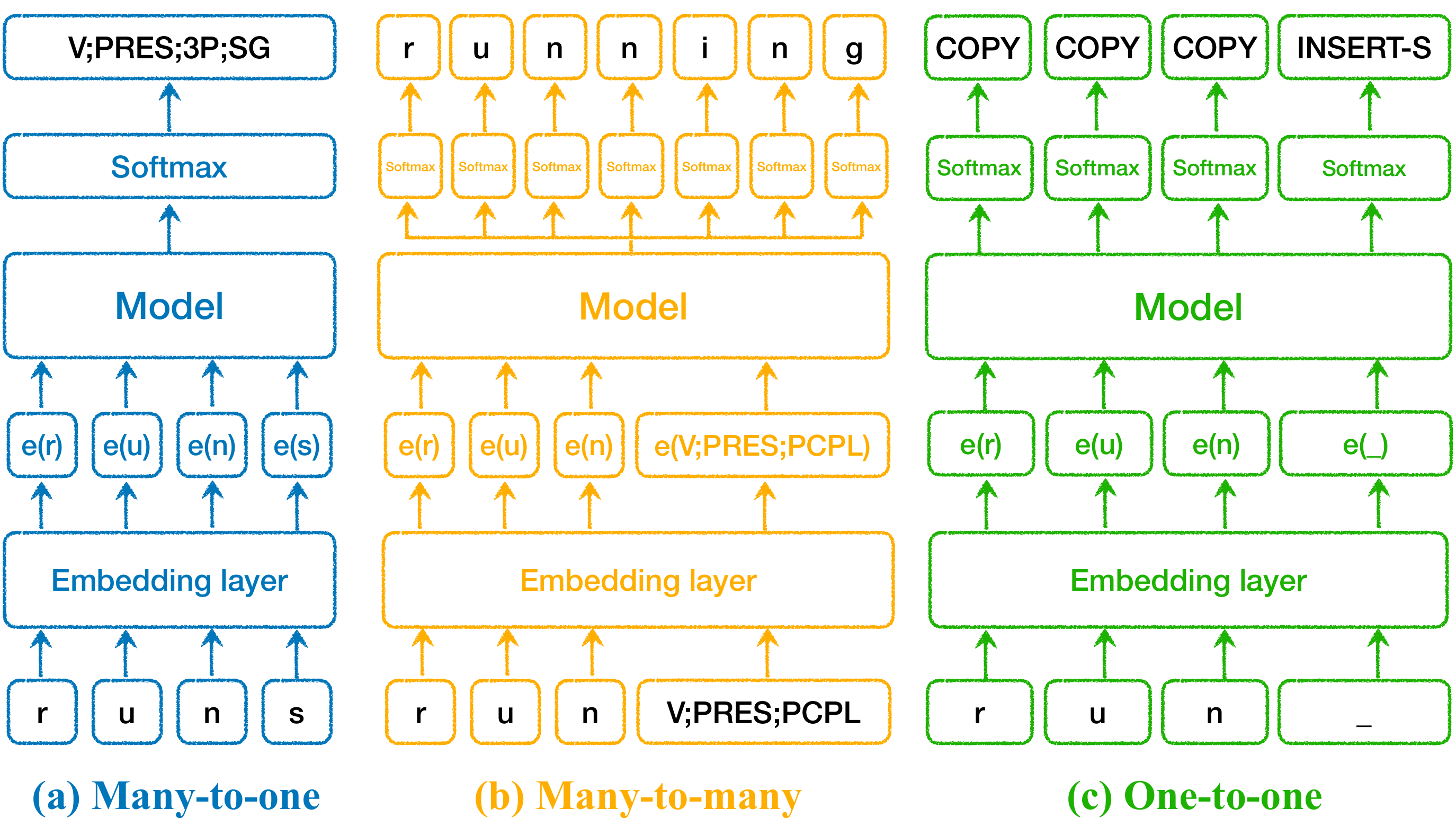}
    \caption{Illustration of three different learning scenarios}
    \label{fig:scenarios}
\end{figure}

In the \textit{many-to-one} scenario, the model first reads in the input text and gets the representation of the text. Then it makes the prediction, which is an output with only one symbol or one chunk of symbols, like POS or MSD tags as a chunk. The typical task for this scenario is \textbf{classification}. Morphological tagging is such a task if we treat each unique MSD tag as a class, as shown in Figure \ref{fig:scenarios}(a). 

In the \textit{many-to-many} scenario, the model first reads in the text, gets the representation of the text and then generates another sequence of symbols which may or may not be of the same length as the input. This scenario is usually referred to as \textbf{sequence-to-sequence transduction}.  This scenario is illustrated in Figure \ref{fig:scenarios}(b), with the morphological inflection task. Other morphological generation tasks can also be naturally dealt with as this scenario.

In the \textit{one-to-one} scenario, the model reads in the input text and outputs a prediction for each corresponding symbol in the input. The typical task for this scenario is \textbf{sequence labeling}. For example, in the morphological inflection task shown in Figure \ref{fig:scenarios}(c) where we need to generate the present participle for the English verb \textit{run}, if we predict the inflected form by making the model to predict the edit actions to convert the input string to the expected output string, the correct operations should be \textit{Copy}, \textit{Copy}, \textit{Copy}, and \textit{Insert-s}.

\subsection{Neural network architectures}

The neural network architectures involved in the conversion process which happens in the model part in the plots in Figure \ref{fig:scenarios}, fall into four main types: feedforward, convolutional neural networks (CNNs), recurrent neural networks (RNNs) and the Transformer. The architectures can be mixed and combined with one another or with non-neural models like HMMs, CRFs, etc. to form more complicated models for different tasks. A softmax layer is usually applied to the output of neural encoding architectures for classification. For sequence transduction tasks, the encoder-decoder structure has been very successful. A lot of progress have been made to improve this structure recently. The last subsection will present the encoder-decoder structure.

\subsubsection{Feedforward}
\label{subsect:feedforward}

Fully connected feedforward neural networks \cite{hornik1989multilayer}, also called multi-layer perceptron, are non-linear and can be conveniently applied to classification problems as well as structured prediction problems. The feedforward neural network architecture has the drawback of requiring fixed length representations. One technique to represent unbounded number of symbols is through the bag of words method. The bag of words method simply adds up vector representations for multiple symbols (and perhaps also divides the sum by the number of symbols) to get the vector representation for the larger unit consisting of those symbols. However, the bag of words method has the problem of discarding order information in the input \cite{goldberg2016primer}.

\subsubsection{CNN}
\label{subsect:cnn}

Convolutional neural networks (CNNs) \cite{lecun1995convolutional}, also called convolution-and-pooling, are designed to pick out local predictors in a large structure and combine them to form a fixed representation of the structured input. Specifically, a non-linear function, called filter, is applied to each instantiation of a \textit{k}-sized sliding window over the input to capture the important information within that window and transform it into a fixed-size representation. Then a pooling operation is conducted to combine the result from different windows into a single representation. Several convolutional layers can be applied in parallel. The most commonly used pooling techniques are max pooling and average pooling. This architecture can generate better vector representations than CBOW \cite{goldberg2016primer}.

\subsubsection{RNN}
\label{subsect:rnn}

Recurrent neural networks (RNNs) are designed to capture the structured information in sequences of arbitrary length, which is especially suitable for natural language data processing. There are two main different types of RNNs: (1) vanilla RNN \cite{elman1990finding}, (2) gated RNN which can be realized with either the Long-Short Term Memory (LSTM) architecture \cite{hochreiter1997long} or the Gated Recurrent Unit (GRU) architecture \cite{cho2014learning}. Vanilla RNN represents the state at postion \textit{i} as a linear combination of the input at this position and the previous state passed through a non-linear activation (usually ReLU or tanh). It is effective to capture sequential information, but is hard to train effectively due to the vanishing gradient problem as pointed out by \newcite{hochreiter2001gradient}. The LSTM architecture and the GRU architecture add ``memory cells'' controlled by gating components to tackle the problem in the vanilla RNN architecture. 

Bidrectional RNNs \cite{schuster1997bidirectional} encode the sequence forward and backward, and combine (usually by concatenation) the output of the forward pass and the backward pass at the same position in the sequence to get the representation for the position at hand, and thus capture both the past and future context \cite{graves2008supervised,goldberg2016primer}. The bottom part of Figure \ref{fig:seq2seq-attn} provides a sketch of the bidirectional RNN architecture.

\subsubsection{Encoder-decoder structure and the Transformer}
\label{subsec:encoder-decoder}

\begin{figure}
    \centering
    \includegraphics[width=0.9\textwidth]{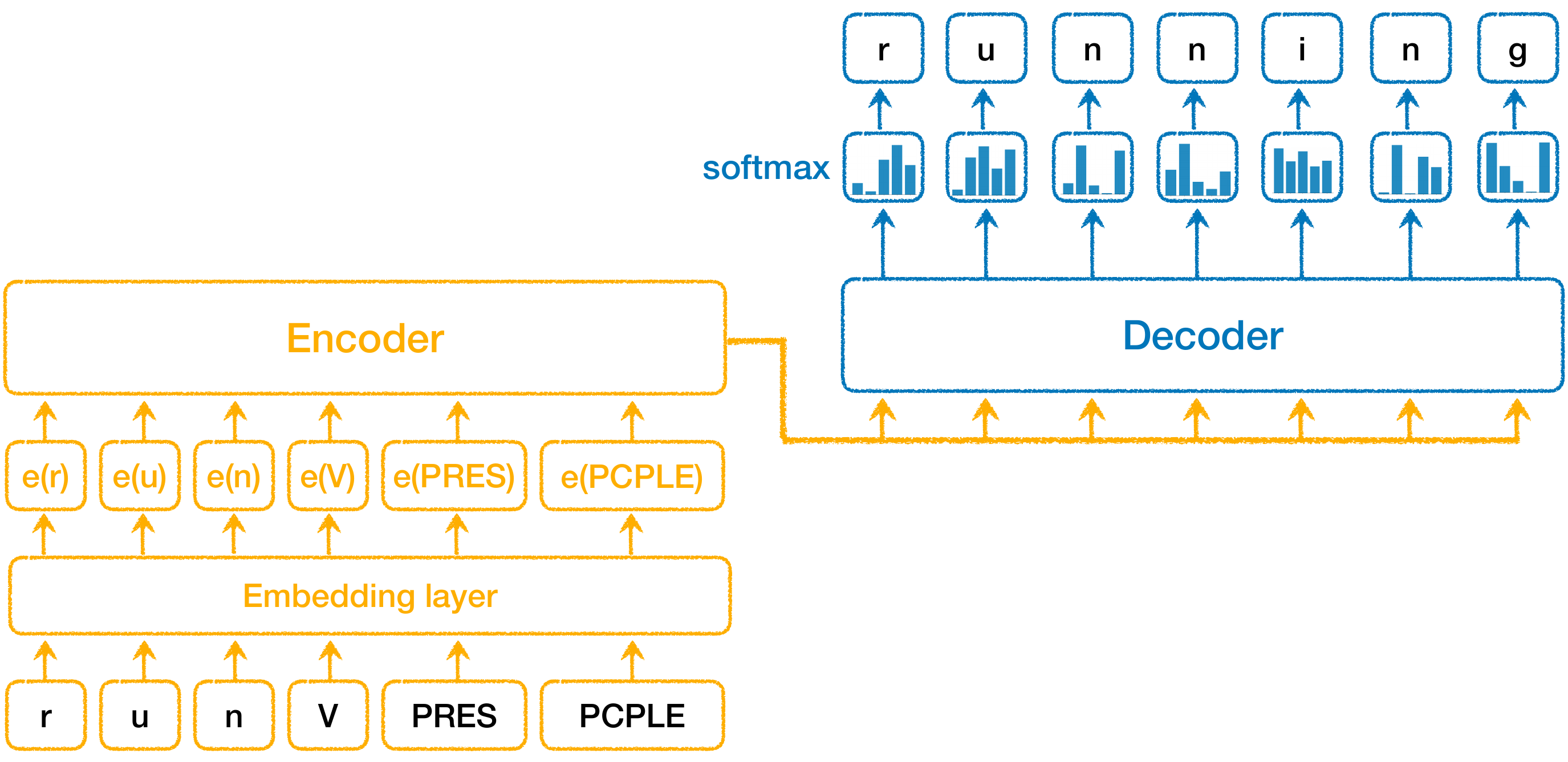}
    \caption{Illustration of the sequence-to-sequence architecture, with the morphological inflection task as an example: Given the English verb lemma \textit{run}, generate its inflected form as the present participle.}
    \label{fig:seq2seq}
\end{figure}

The encoder-decoder structure \cite{cho2014properties,sutskever2014sequence}, also called sequence-to-sequence architecture (or ``seq2seq'' for short), was initially designed for machine translation and then successfully applied to more generally sequence transduction problems. The encoder-decoder structure, illustrated in Figure \ref{fig:seq2seq}, basically breaks down the input-to-output conversion process into encoder transformations and decoder transformations. The inner architecture of the encoder and decoder transformations can be formed from any or any combination of the three basic structures introduced above. In a vanilla seq2seq architecture, the encoder converts the input sequence to a vector representation. The same vector representation is passed as input, usually together with other information based on the task, to the decoder at each time step to generate the output sequence. The vanilla architecture forces the encoded input vector representation to contain all the information from the input sequence for decoding and the decoder to be able to extract that information from the fixed-length representation from the encoder output. This structure works well in general, but the attention mechanism \cite{bahdanau2015} relieves the burden of the encoder and improves the model for sequence transduction. 

\begin{wrapfigure}{R}{0.5\textwidth}
    \includegraphics[width=0.4\textwidth]{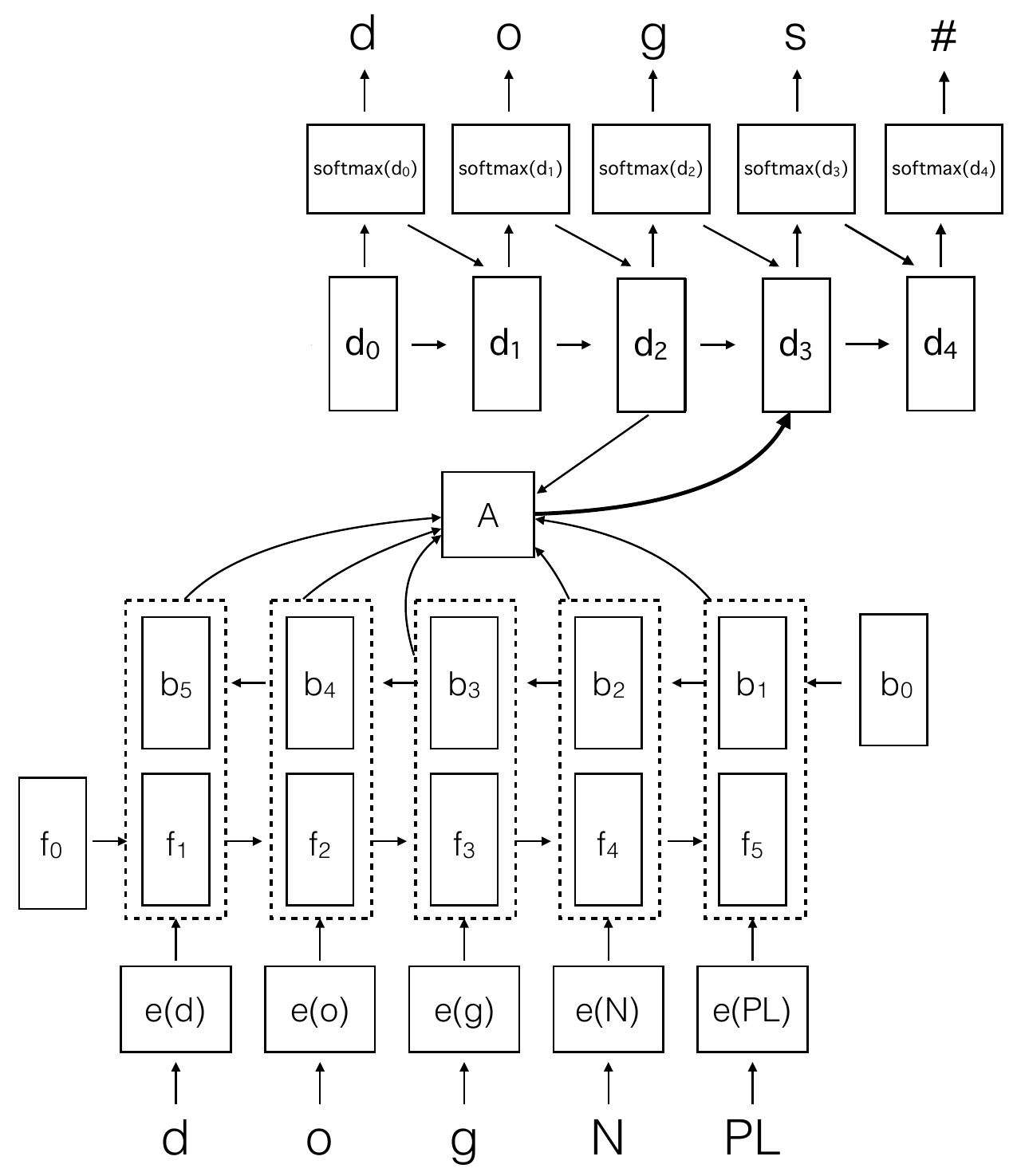}
    \caption{Illustration of the seq2seq architecture with an attention mechanism. The encoder illustrated here (bottom part) is bidirectional. The attention illustrated in the figure is that for generating the character \textit{s}. The figure was created by reference to \protect\cite{silfverberg-etal-2017-data}. }
    \label{fig:seq2seq-attn}
\end{wrapfigure}

The attention mechanism learns a weight vector applied to every encoding state for each decoding state. Figure \ref{fig:seq2seq-attn} illustrates a seq2seq architecture with a bidirectional encoder, an attention mechanism and a decoder. The soft attention mechanism calculates the context vector as weighted sum of the encoder hidden states, while the hard attention mechanism selects certain encoder hidden states to use \cite{luong2015multi}. 

The RNN-based encoder-decoder architecture with attention has been very successful with string transduction. However, the sequential processing of the RNN is problematic: During the sequential processing, information may deteriorate when passing through the states over time. In addition, due to sequential processing, parallel computation is difficult to apply, making the architecture too slow for large scale use. The Transformer model \cite{vaswani-et-al2017attention} attempts to improve the seq2seq model to overcome its reliance on sequential processing. The Transformer still uses the encoder-decoder structure in its architecture, but it is stateless and does not use the RNN architecture. Instead, it uses multi-head attention to compute the input and output representations with self-attention layers as well as the combination of the input. It makes predictions with ``encoder-decoder attention'' layers which mimics the typical attention mechanism in previous seq2seq models.  

\section{Type-based computational morphology}
\label{sec:type-based}

Type-based computational morphology refers to the computational processing of words out of context, like words in lexicon. The word form is provided in isolation, sometimes together with labels about morphosyntactic information.


\begin{wrapfigure}{R}{0.45\textwidth}
\centering
\includegraphics[width=0.45\textwidth]{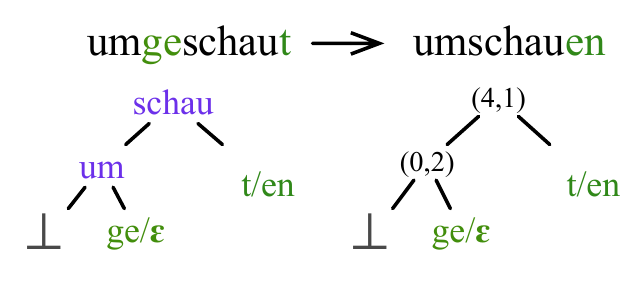}
\caption{Edit tree to transform the German word \textit{umgeschaut} ``looked around'' to its lemma \textit{umschauen} ``to look around''. The blue nodes in the tree on the left are common substrings between the inflected form and its lemma form. These nodes are replaced with the span information in the edit tree on the right for generalization. \textit{ge}/$\varepsilon$ means deleting \textit{ge}, \textit{t/en} means substitute \textit{t} with \textit{en}, $\bot$ means no edit action. This example is adopted from \protect\newcite{muller2015joint}}
\label{fig:edit-tree}
\end{wrapfigure}

\subsection{Morphological generation with neural models -- (re)inflection}
\label{subsec:model_and_inflection}

\begin{figure}
    \centering
    \includegraphics[width=0.7\textwidth]{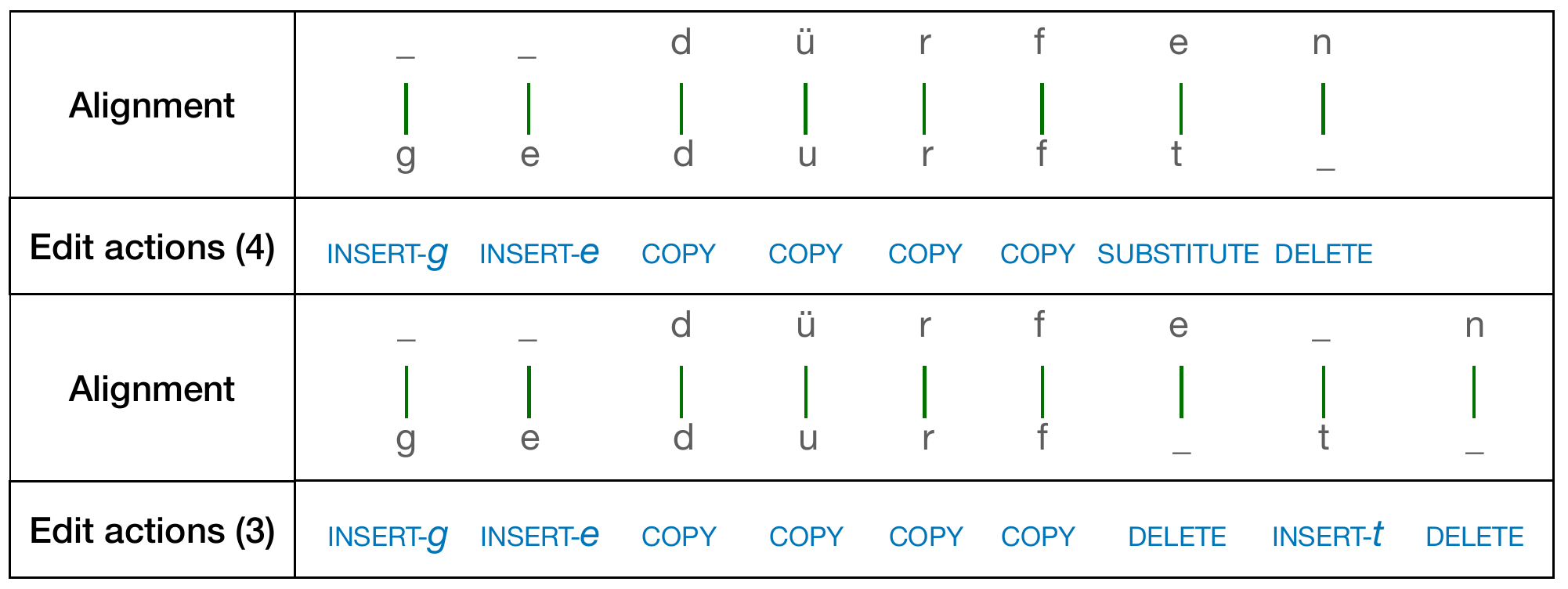}
    \caption{Alignment and edit action example for the German verb \textit{dürfen} ``to be allowed'' and its past participle form \textit{gedurft} ``have/has/had been allowed''. Rows 1 and 2 show the alignment with four distinct edit actions. Rows 3 and 4 show the alignment with three distinct edit actions.}
    \label{fig:align}
\end{figure}

Morphological generation, including inflection, reinflection, etc, belongs to the type of problem of mapping from one sequence to another. Before neural network models were applied, this type of problem has traditionally ``been tackled by a combination of hand-crafted features, alignment models, segmentation heuristics, and language models, all of which are tuned separately.'' \cite{yu2016online}(p1307) Alignment and edit actions are still commonly leveraged in many neural network approaches \cite{aharoni2016improving,aharoni2017morphological,makarov2018conll,ribeiro-etal-2018-local,makarov2018neural,cotterell2017graphical,kann2017sharedtask,chakrabarty2017context,kondratyuk-straka-2019-75}. Edit actions usually include \texttt{COPY}, \texttt{DELETE}, \texttt{INSERT} and \texttt{SUBSTITUTE}. The \texttt{SUBSTITUTE} edit action can be treated as \texttt{DELETE} and \texttt{INSERT}. Figure \ref{fig:align} provides examples for the alignment with four and three distinct edit actions respectively.

\newcite{faruqui2016morphological} is the first work to apply the \textbf{encoder-decoder} model to \textbf{inflection} generation.
They use a bidirectional LSTM for the encoder, which takes the character sequence of the lemma form as input. The output of the encoder is fed into the LSTM decoder at each time step together with the decoder output from the previous time step. Their decoder does not use the attention mechanism. The authors experimented with learning separate models for each inflection type, or learning a shared encoder on the full inflection table and training a separate decoder for each inflection type. They also experimented with improving the model by using \textbf{unlabeled data}, i.e. to learn a language model over character sequences in a vocabulary and use the language model to rerank the decoding output by beam search or to interpolate the probability of the next word predicted by the language model with the probability of the next word predicted by the decoder. They report results obtained using \textbf{an ensemble of models} on the dataset created by \newcite{durrett2013supervised} and extended by \newcite{nicolai2015inflection}, including full inflection tables for German, Finnish, Spanish, Dutch, and French in this dataset. Their model achieved results better than or comparable to previous state-of-the-art models on the same dataset \cite{durrett2013supervised,ahlberg2014,ahlberg2015,nicolai2015inflection}. They also compare their model with the encoder-decoder model and the attention-based encoder-decoder model which doesn't incorporate the encoder output at each time step of encoding. They found that their model achieves the best performance. There are two main problems with \newcite{faruqui2016morphological}'s models resulting from their design of training a separate model for each distinct MSD tag. First, they have to train as many models as there are MSD tags in the language, which will be cumbersome if the language has a lot of inflection types. Second, they need a lot of labeled data: no information can be shared between between different inflection types though similar rules may apply to the inflection for several MSD tags, and consequently, they need a lot of labeled data for each MSD tag in order to train good models. In addition, partially filled paradigms can't be utilized with their approach due to its reliance on complete inflection tables.

The shared task of SIGMORPHON in 2016 is about morphological (re)inflection, where morphological datasets of 10 languages with diverse typological characteristics were introduced. In \textbf{a labeled inflection task}, what's given is the lemma and a target tag, and the task is to predict the inflected form for the lemma corresponding to the target tag, i.e. \textit{lemma + target MSD} $\rightarrow$ \textit{target form}. In \textbf{a labeled reinflection task}, what's given is a source form, a source tag and a target tag, and the task is to predict the inflected form for the word corresponding to the target tag, i.e. \textit{source form + source MSD + target MST} $\rightarrow$ \textit{target form}. There is also \textbf{a more difficult version of the labeled reinflection task} which does not provide the MSD tag for the source form, i.e. \textit{source form + target MSD} $\rightarrow$ \textit{target form}. Nine teams submitted systems for this shared task. In general, the three teams who employed neural network models \cite{kann2016med,aharoni2016improving,ostling2016morphological} outperformed by a large margin other teams who employed non-neural approaches of alignment and transduction pipelines or reduction of the problem to multi-class classifications with linguistic-inspired heuristics \cite{alegria-etxeberria-2016-ehu,nicolai-etal-2016-morphological,liu-mao-2016-morphological,king-2016-evaluating,sorokin-2016-using,taji-etal-2016-columbia}. The winning system \cite{kann2016med} uses the same approach as \newcite{kann2016single}, and applies the sequence-to-sequence architecture initially developed for machine translation successfully to the morphological (re)inflection task. Specifically, as illustrated in Figure \ref{fig:seq2seq-attn}, they use a GRU-based \cite{cho2014properties} bidirectional encoder-decoder model \cite{sutskever2014sequence} with a soft attention mechanism \cite{bahdanau2015}, which takes as input the character sequence in the input word along with features in the source and target tags and predicts the target inflected form character by character. This formatting of the data is the key innovation. They used an ensemble of 5 models and selected the predicted form with the majority vote from the models as the final prediction. Different from \newcite{faruqui2016morphological}, they trained one model per language by treating each feature in the MSD tag as a separate symbol, and thus the approach has less strict requirement about the data. This system outperformed other models submitted for the same shared task by a large margin and pushed the morphological generation performance to a higher level across languages. Afterwards, the \textbf{bidirectional encoder-decoder with an attention mechanism} architecture has been widely applied to morphological generation tasks, and tweaks have been made to further improve the model. \newcite{kann2016single} propose the Perfect Observed Edit Trees (POET) method for correcting morphological reinflection predictions. An edit tree example is provided in Figure \ref{fig:edit-tree} on page \pageref{fig:edit-tree}. They evaluate combining this method with the neural model in \newcite{kann2016med}. They find that adding the POET method can improve the neural model and outperform \newcite{faruqui2016morphological}'s model by an even larger margin. However, the POET method relies on a large enough training dataset to cover the possible \textbf{edit trees} in the language.

The second-ranked neural model team \cite{aharoni2016improving} in SIGMORPHON 2016 shared task submitted two systems: One system is an \textbf{encoder-decoder} model with the encoder augmented with a bidirectional LSTM, and trains a network for each part-of-speech type. 
The model also has template-inspired components to align the training data first and train the network to either predict or copy a character at a given position. The other system from this team is based on neural discriminative string transduction, i.e. to first align the training data and get the sequence of \textbf{edit actions} to derive the inflected form from the source form, and then use the encoder-decoder model to predict the sequence of edit actions given the input sequence and the set of target MSD features, a hard-attention mechanism which has been tweaked more in later works like \newcite{aharoni2017morphological}. 

The third-ranked system \cite{ostling2016morphological} also uses a seq2seq model, but without an attention mechanism. They used multi-layers of LSTM for the decoder and train one model for each language to enable better parameter sharing. 
In addition to using the CNN architecture to extract features which are the inputs to the encoder-decoder layers of the LSTM architecture together with other feature vectors (e.g. character embedding, MSD vectors), \newcite{ostling2016morphological} also experimented with using only the 1-dimensional residual network architecture \cite{he2016identity} for encoding. They found that the accuracy is consistently improved by using convolutional layers together with LSTM for encoding and that purely convolutional encoding achieves even higher accuracy in many cases.



The vanilla encoder-decoder model has the limitation that decoding can only start after the whole input sequence has been encoded, and attention weights are treated as output of a deterministic function. In order to overcome this bottleneck, \newcite{yu2016online} propose an online neural sequence-to-sequence model where the encoder is a unidirectional LSTM and beam search is used to marginalize a sequential latent variable for alignment in the decoding process. This allows the model to decide how much to encode before decoding each part. The model was evaluated on the tasks of abstractive sentence summarization and morphological inflection. The same dataset as \newcite{faruqui2016morphological} was used for the morphological inflection experiments, and the model was trained separately for each type of inflection, the same setting as the factored model by \newcite{faruqui2016morphological}. It achieved comparable results as previous neural and non-neural state-of-the-art models \cite{faruqui2016morphological,durrett2013supervised,nicolai2015inflection} and the better performance on the Finnish data, whose stems and inflected forms are the longest, supports the advantage of this model. However, this model has not been applied to more tasks in computational morphology processing. One reason may be that the problem this model targets at, i.e. enabling the model to generate without waiting for having the entire input sequence encoded, is not a serious problem for morphology since most words are not very long. In addition, the Transformer model \cite{vaswani-et-al2017attention} overcomes the problem caused by very long sequences and produces the current state-of-the-art performance on the morphological inflection task \cite{vylomova-etal-2020-sigmorphon}.

\begin{figure}
    \centering
    \includegraphics[width=\textwidth]{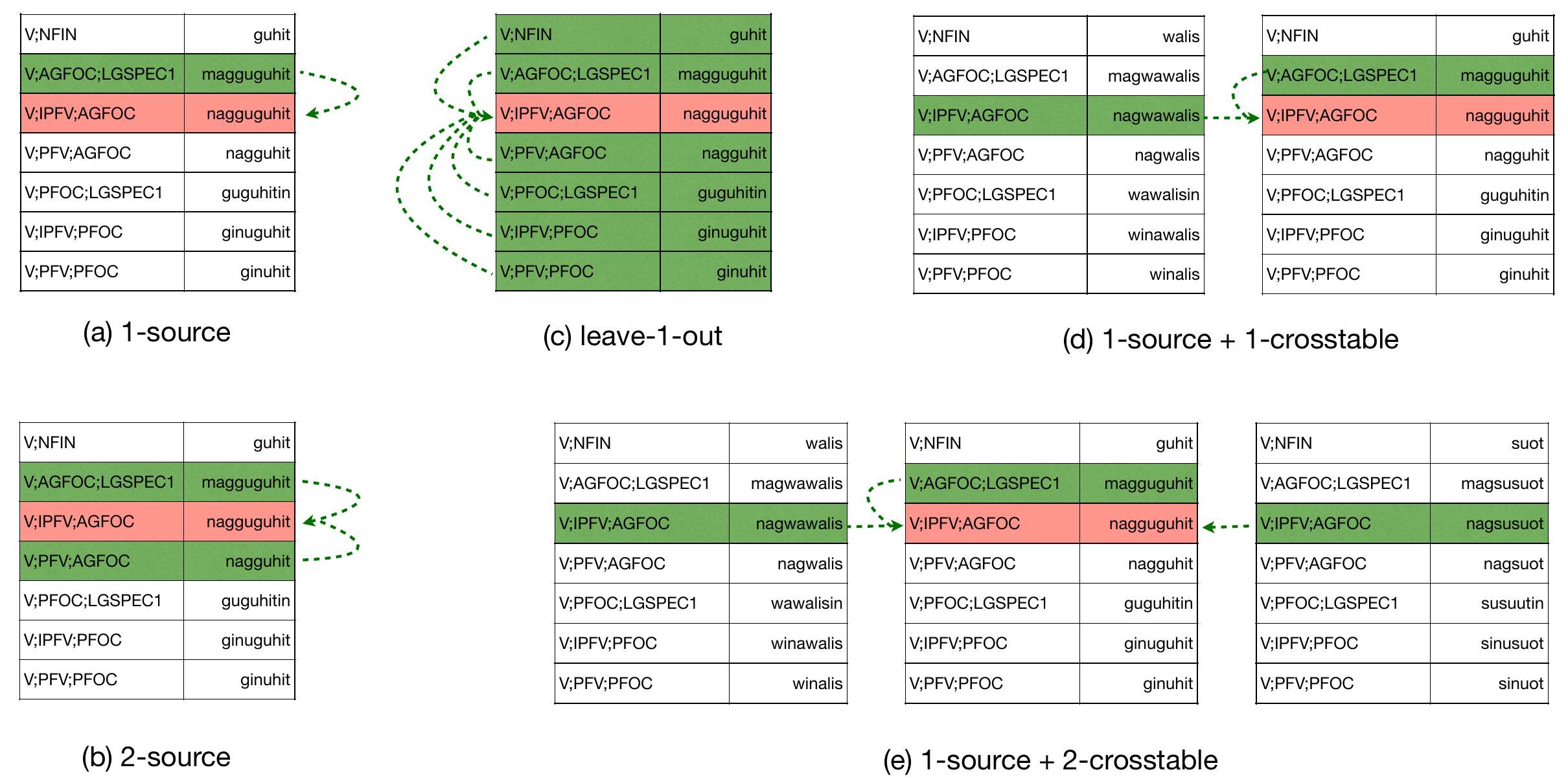}
    \caption{Illustration of multi-source (re)inflection and cross-table (re)inflection.  Green cells are the source. Red cells are the target. The figure is adopted from \protect\newcite{liu-hulden-2020-analogy}}
    \label{fig:multi-source}
\end{figure}

\newcite{kann-etal-2017-neural} extend the typical \textbf{morphological (re)inflection} and use \textbf{multiple source} form-tag pairs to predict the inflected form for the target tag. Figure \ref{fig:multi-source}(a) illustrates the typical reinflection setup where a single source form is used to predict the target form. Figure \ref{fig:multi-source}(b) illustrates the case where two source form-tag pairs are used for predicting a target slot. Figure \ref{fig:multi-source}(c) illustrates the case where multiple source form-target pairs are used. \newcite{kann-etal-2017-neural}'s model is still the bidirectional encoder-decoder model with an attention mechanism, but uses one bidirectional encoder to encode each pair of source form and source tag for a lemma, and the decoder takes as input the attention-weighted sum of the encoder states. They use \textbf{full inflection tables}, and their experiments show that such a multi-source approach outperforms the typical single-source approach, especially when the multiple encoders share parameters. They also experimented with using multiple source form-tag pairs by concatenating all sources and encoding the concatenated form-tag pairs by one encoder. Their results indicate that the two ways of using multiple source form-tag pairs achieve comparable results.

The Transformer model was applied to the morphological (re)inflection task by \newcite{wu2020applying}. In SIGMORPHON 2020 shared task 0 on morphological inflection, the Transformer model is the model architecture which produces the best performance \cite{vylomova-etal-2020-sigmorphon,liu-hulden-2020-leveraging}. \newcite{moeller-etal-2020-igt2p} compared the performance of the Transformer model \cite{vaswani-et-al2017attention} and the seq2seq model with exact hard monotonic attention \cite{wu-cotterell-2019-exact} for inflection of noisy data in low-resource scenario. Their results support the advantage of the Transformer model as well.

\subsubsection{Low-resource scenario}


Though the performance of the seq2seq model is very impressive on the morphological (re)inflection task when there are abundant labeled data, the performance in the low-resource scenario drops dramatically. In order to improve the performance of neural network models in low-resource scenarios, efforts have be made to customize the model architecture for morphology tasks or to augment the training data by leveraging unlabeled data, data hallucination, labeled data from other languages etc.

Both machine translation and morphological (re)inflection are essentially string-to-string transduction tasks. The difference is that in morphological (re)inflection, a large part of the output is usually a copy of the input. Inspired by this difference, \newcite{aharoni2017morphological} present an encoder-decoder model with a hard-attention mechanism, which further improves the second system they created for SIGMORPHON 2016 shared task \cite{aharoni2016improving}. The encoder is a bi-directional RNN which takes as input the concatenation of a forward RNN and a backward RNN states over the word's characters, and at decoding time, a control mechanism attends to a single input state and decides whether to write a symbol to the output sequence or advance the attention pointer to the next state from the bi-directionally encoded sequence. This model produces state-of-the-art results compared to previous neural and non-neural approaches. Its advantage is especially outstanding in the low-resource scenario. In addition, it's computationally more efficient than the soft attention mechanism for decoding: the model with hard monotonic attention allows linear decoding time while the decoding time for the soft attention mechanism is quadratic with regard to sequence length. This model has been widely adopted for low-resource languages, such as \newcite{makarov2017align} who develop the winning systems with this architecture for CoNLL-SIGMORPHON 2017 shared task on morphological (re)inflection, \newcite{cotterell2017graphical} who apply the model to generate inflected word forms for principal parts morphological paradigm completion, and \newcite{junczys-dowmunt-grundkiewicz-2017-exploration} who apply the model to machine translation and automatic post-editing. However, this model is not truly end-to-end because it has a pipeline part: it relies on an external aligner to align the input string with the output string before training.

\newcite{makarov2018neural} explain the hard attention mechanism as a neural transition-based system over edit actions and highlight the similarity of this approach with traditional weighted finite-state transducer (WFST) approaches. In addition, they replace the maximum likelihood estimation (MLE) training in \newcite{aharoni2017morphological} with \textbf{minimum risk training} \cite{och-2003-minimum,smith-eisner-2006-minimum,shen-etal-2016-minimum}, making the training process truly end to end. The model was evaluated on morphological inflection, reinflection and lemmatization tasks, and generated results comparable to or better than the state of the art. This is also the architecture used in the winning system for the first subtask in CoNLL-SIGMORPHON 2018 shared task. \newcite{makarov2018imitation} apply \textbf{imitation learning} to train neural transition-based approach and achieved similar effects. 

\newcite{wu-etal-2018-hard} use the hard non-monotonic attention with the seq2seq model. Their evaluation of the model on CoNLL-SIGMORPHON 2017 shared task 1 shows that this model achieves better results than the encoder-decoder with soft attention model in the high training data setting. \newcite{wu-cotterell-2019-exact} generalize the architecture of \newcite{wu-etal-2018-hard} and propose \textbf{the exact hard monotonic attention mechanism for string transduction at the character level}, which enforces strict monotonicity, i.e. to attend to exactly one vector at memory at each output time step, and learns a latent alignment and transduction jointly. This seq2seq model with exact hard monotonic attention achieved state-of-the-art performance for CoNLL-SIGMORPHON 2017 shared task 1 at the high training data setting. 

\newcite{ribeiro-etal-2018-local} thoroughly apply the idea of converting the morphological generation problem to a sequence labeling task by predicting edit operations, and present a technique for this transformation. They demonstrate that this approach performs comparable to the state of the art \cite{kann2016med,aharoni2017morphological} for morphological inflection. 

\newcite{zhou2017multi} propose the \textbf{multi-space encoder-decoder model} for labeled sequence transduction. Compared to the common encoder-decoder architecture with an attention mechanism, their model features an intermediate step, the multi-space variational auto-encoder, which is a generative model that can process both continuous and discrete latent variables and enables the model to take advantage of both \textbf{labeled and unlabeled data} effectively. It was evaluated on the task of morphological reinflection, specifically the data for the 3rd subtask (i.e. $source form + target MSD \rightarrow target form$) of SIGMORPHON 2016 shared task, and achieved much better performance on most languages than the MED system \cite{kann2016med} without model ensembling. When unlabeled data is used to train the model in a \textbf{semi-supervised} way through the multi-space variational auto-encoder, the result surpasses the MED system without model ensembling for all languages except Spanish. \newcite{kann-schutze-2017-unlabeled} is another work that found \textbf{unlabeled} data helpful. 
They take the \textbf{multi-task learning} approach by making the model to jointly predict the target inflected word form and map the unlabeled data with a special tag to itself, a task which they named as autoencoding. The use of unlabeled data improved the accuracy for the 8 languages they experimented with.

CoNLL-SIGMORPHON 2017 shared task 1 \cite{cotterell2017conll} is on universal morphological inflection for 52 languages, which features 3 data-size settings for most languages: low (100 lemma-target MSD-target form triples for training), medium (1,000 lemma-target MSD-target form triples for training), and high (10,000 lemma-target MSD-target form triples for training) settings. There were 24 systems submitted for this subtask, all but one of which have a neural component, specifically the LSTM or GRU encoder-decoder architecture with an attention mechanism and all of which focus on the \textbf{low-resource} problem by model improvements like incorporating the model with the appropriate inductive bias, or \textbf{data augmentation} techniques like making use of unlabeled data or dummy data. The purpose of data augmentation is to increase the amount of training data and help the model learn the appropriate copy bias. The copy bias is critical in morphological (re)inflection because different from other sequence transduction tasks like machine translation, the transduction from the input form to the target inflected form usually involves copying the input symbol at many places. 
\newcite{bergmanis2017training}'s system is the overall winning system for the first subtask at high resource setting. In addition, they experimented with autoencoding of unlabeled data or random strings to make the model bias towards copying, as well as using labeled data in other languages to help training models for the current language (i.e. \textbf{cross-lingual transfer}). They found that autoencoding of random strings can improve the performance almost as well as autoencoding unlabeled words in both high and medium settings. What they do with autoencoding is to create additional training pairs by making an input string with a special tag like \textit{COPY} to predict the input string itself. \newcite{zhou-neubig-2017-morphological} use \textbf{unlabeled data} with a variational autoencoder and latent variables \cite{zhou2017multi}. \newcite{nicolai-etal-2017-cant} try to augment the data by training a tagger with the training data and generate more labeled data by tagging more unlabeled data. \newcite{silfverberg-etal-2017-data} and \newcite{kann2017sharedtask} augment the data by creating artificial training data. 
Most systems use the soft attention mechanism, but the winning team \cite{makarov2017align} develop systems based on the \textbf{hard monotonic attention mechanism} \cite{aharoni2016improving,aharoni2017morphological}. Specifically, their first system extends the hard monotonic attention model with a copy mechanism and their second system is a neural state-transition model which learns to copy explicitly. \newcite{nicolai-etal-2017-cant} use hard monotonic alignment by combining a discriminative string transduction model with the neural approach, though the system did not perform very well. 
The results of different teams indicate that the encoder-decoder architecture with an attention mechanism can handle morphological inflection well even when the training data is limited and that different model biases and data augmentation techniques compliment each other. 

\begin{figure}
    \centering
    \includegraphics[width=\textwidth]{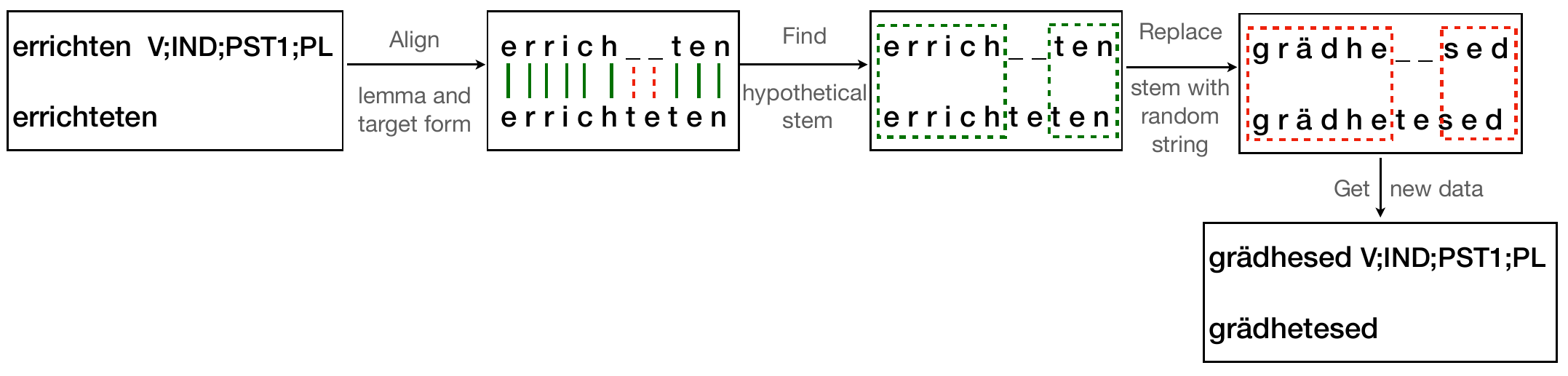}
    \caption{Illustration of the data hallucination method by \protect\newcite{anastasopoulos-neubig-2019-pushing}. The only difference in \protect\newcite{silfverberg-etal-2017-data}'s method is \protect\newcite{silfverberg-etal-2017-data} take the longest common substring as the hypothetical stem, i.e. \textit{errich} in this example.}
    \label{fig:data_hallucination}
\end{figure}

The data hallucination method by \newcite{silfverberg-etal-2017-data} is very effective for augmenting the training data in the low-resource scenario. This method was improved by \newcite{anastasopoulos-neubig-2019-pushing} to take into consideration discontinuous stems as well. To be specific, both methods create hallucinated data following the steps illustrated in Figure \ref{fig:data_hallucination} --- First align the lemma with the inflected form, then find the hypothetical stem part, next generate a random string by uniformly sampling from the alphabet as dummy stem and replace the hypothetical stem with the dummy stem to form dummy training triples. The new dummy triple are added to the training set for data augmentation. The difference between \newcite{anastasopoulos-neubig-2019-pushing} and \newcite{silfverberg-etal-2017-data} is in how the hypothetical stem is defined: \newcite{silfverberg-etal-2017-data} take the longest common substring between the source form and the target form as the stem while \newcite{anastasopoulos-neubig-2019-pushing} take all the common substrings between the source form and the target form as the stem and thus include the discontinuous stem as well.

The first subtask of CoNLL-SIGMORPHON 2018 shared task \cite{cotterell2018conll} is the same as that for the previous year: given a lemma form and a target MSD tag, the system is expected to predict the target inflected form in low, medium, high data conditions, and the system should be language-agnostic. The number of languages increased to 103 from 52. 27 systems were submitted for this shared task. Nearly all systems have a neural component, the same as in 2017. Though based on highly-ranked systems from previous years' shared tasks, further tweaks were made to the systems and 41 languages in the 52 included in 2017 shared task got better results in the low-resource setting. In order to deal with the \textbf{low-resource setting}, systems this year still try to learn \textbf{sequences of edit operations} or guide the neural model toward \textbf{copy bias} by creating artificial training data. Ensembling is also used by most systems to improve performance. The winning system \cite{makarov2018conll} uses the fully end-to-end transition-based model by \newcite{makarov2018neural}. 
\newcite{rama-coltekin-2018-tubingen} develop a multilingual multi-task seq2seq model, but find that the model didn't surpass the baseline in high and medium data settings.

Character-level \textbf{language modeling} has been used to augment neural models for morphological (re)inflection. For example, \newcite{nicolai-etal-2018-string} leverage target language models derived from unannotated target corpora and make further improvements over the winning system for CoNLL-SIGMORPHON 2017 shared task 1 at low data setting (i.e. with 100 \textit{lemma-tag-form} triples for training). \newcite{sorokin-2018-gain} also attempt to improve the neural inflection model with character-level language modeling, but they only see marginal improvements.

Inspired by the analogy mechanism for inflecting unknown words by reference to known words, \newcite{liu-hulden-2020-analogy} propose to add inflected forms for the target slot in other inflection tables where this slot is given to the input source forms to predict the target form. This data manipulation method improves the performance of the Transformer model in extremely low-resource situations. Figures \ref{fig:multi-source} (d) and (e) illustrate the cross-table data manipulation method. Though the interparadigmatic and intraparadigmatic aspects of analogy has been leveraged before neural network approaches have been applied to morphology \cite{ahlberg2014,hulden2014,liu-hulden-2017-evaluation,silfverberg-etal-2018-computational}, \newcite{liu-hulden-2020-analogy} is the first attempt to explicitly provide both interparadigmatic and intraparadigmatic analogy sources for neural network models to learn from, and other existing work provide only the intraparadigmatic analogy source, leaving it for the model to infer the interparadigmatic analogical information implicitly.

\begin{table}
\centering{
\begin{adjustbox}{width=0.85\textwidth}
\begin{tabular}{lll|lll}
\toprule 
\multicolumn{3}{c|}{\textbf{Spanish}} & \multicolumn{3}{c}{\textbf{Portuguese}} \\
\midrule
comer & \textbf{comemos} & V;IND;PRS;1;PL & comer & \textbf{comemos} & V;IND;PRS;1;PL \\
comer & coméis & V;IND;PRS;2;PL & comer & comeis & V;IND;PRS;2;PL \\
comer & comen & V;IND;PRS;3;PL & comer & comem & V;IND;PRS;3;PL; \\
comer & \textbf{comeríamos} & V;COND;1;PL & comer & \textbf{comeríamos} & V;COND;1;PL \\
comer & comeríais & V;COND;2;PL & comer & comeríeis	& V;COND;2;PL \\
... & ... & ... & ... & ... & ... \\
\bottomrule
\end{tabular}
\end{adjustbox}
\caption{A fraction of UniMorph inflection tables for \textit{comer}, a verb used in both Spanish and Portuguese, meaning ``to eat''. Identical inflected forms are highlight in bold.}
\label{tab:language-similar}
}
\end{table}

\subsubsection{Cross-lingual transfer}

Languages share features, some related languages also share words and the inflection mechanisms are common across language groups. For example, Spanish and Portuguese are two languages that are closely related to each other. They shared words, features, and inflection mechanisms. Table \ref{tab:language-similar} provides a fraction of UniMorph \cite{kirov2018} paradigms for \textit{comer} ``to eat'' which is used in both Spanish and Portuguese, where we see much similarity. Could neural network models take advantage of the commonality across languages when multilingual resources are available? 

\newcite{bergmanis2017training} experimented with \textbf{cross-lingual transfer} in their system for CoNLL-SIGMORPHON 2017 shared task, but did not find multilingual resources more helpful than random strings. \newcite{rama-coltekin-2018-tubingen} develop a multilingual multi-task seq2seq model for the first subtask of CoNLL-SIGMORPHON 2018 shared task, but the result of their system did not surpass the baseline in medium and high training data settings. The advantage of their system at the low-resource setting is not obvious either. 

The first subtask of CoNLL-SIGMORPHON 2019 shared task \cite{mccarthy-etal-2019-sigmorphon} is about crosslingual transfer for inflection generation. Five teams submitted systems for this subtask. Three teams \cite{coltekin-2019-cross,hauer-etal-2019-cognate,madsack-weissgraeber-2019-ax} did not find obvious improvements in their systems by adding in cross-lingual data. \newcite{anastasopoulos-neubig-2019-pushing} submitted the overall winning system output with the highest average accuracy and third-ranked average Levenshtein distance. Their system encodes the lemma and MSD tags separately: a bidirectional RNN is used to encode the lemma and a self-attention encoder \cite{vaswani-et-al2017attention} is used to encode the tags, after which two attention weight matrices are used to transform the representations into two sequences of context vectors, which are used by the recurrent decoder to generate the output in a two-step process. The two-step attention architecture for the decoder is novel and outperforms the baseline slightly.  The training process of \newcite{anastasopoulos-neubig-2019-pushing}'s morphological inflection method consists of 3 stages: the first stage trains the model to copy, i.e. be monotonic; the second stage is the main training phase when the model is train with both low- and high- resource language data as well as the dummy data created by the data hallucination method; and the last stage fine-tunes the model. They found that the relatedness between languages is closely correlated with the degree of transfer, and the same writing system is also critical for the cross-lingual transfer. These findings are similar to the findings in \newcite{kann2017one-shot} and \newcite{jin2017exploring} who found that the degree of cross-lingual transfer for morphological inflection is strongly influenced by the relatedness between languages though Arabic data can also be leveraged to improve Spanish models. \newcite{anastasopoulos-neubig-2019-pushing} also found that using data from multiple related languages can contribute even more to the low-resource language model. \newcite{peters-martins-2019-ist} submitted the overall second and third best systems in accuracy, which perform the best as to Levenshtein distance. Their models use sparse sequence-to-sequence modeling and attend to lemma and inflection tags respectively with \textbf{a two-headed attention mechanism} \cite{acs-2018-bme}. In addition, their model is more interpretable. However, I think the better performance of their systems over the baseline indicates the advantage of their model, but does not provide enough support for crosslingual transfer effect in morphological inflection and can not exclude the possibility that the cross-lingual data only helped the model get a better copy bias.

\subsection{Morphological generation with neural models -- paradigm cell filling}
\label{subsec:model_and_paradigm}

Closely related to the morphological (re)inflection task is the partial paradigm cell filling problem (see Figure \ref{fig:pcfp} on page \pageref{fig:pcfp} for illustration), as explained in section \ref{subsubsec:pcfp}. When there are more than one slot available in partial paradigms, this task can be easily converted to the morphological (re)inflection task by taking given slots as the source and the missing slots as the target, following the data manipulation methods shown in Figure \ref{fig:multi-source} on page \pageref{fig:multi-source}, and all the techniques for improving morphological (re)inflection can then be applied. However, when there is only one slot given in each partially filled paradigm, the task is much more challenging \cite{malouf2016,malouf2017,silfverberg-hulden-2018-encoder}.

\newcite{malouf2016} and \newcite{malouf2017} apply the neural network approach to the paradigm cell filling task. They utilize neural models but their model does not have the encoder-decoder architecture. What they do is to apply the idea of paradigm function, which maps a \textbf{lexeme} and an MSD tag to a target word form \cite{stump2001inflectional}, and model the paradigm cell filling problem with an \textbf{LSTM generator}. At each time step the model takes as input a lexeme, an MSD tag set, and a partial predicted word form, where the lexeme and each MSD tag are represented as a two-hot vector \mans{This could use an image.} 
which gets concatenated with the one-hot vector representation of the current character after dimension reduction, and output a probability distribution over the next character in the target word form. The probability of the target word form is the product of the probabilities for each predicted character for that target word form. 
They experimented on seven languages of different morphological complexity, having paradigms with 10\% or 40\% random slots missing and reported results on 10-fold validation. They found that their model can tackle the paradigm cell filling problem very well, and tried to interpret their results with linguistic reason and implications. The advantage of \newcite{malouf2016} and \newcite{malouf2017}'s model is that it represents the lexeme, which can be considered as the abstract underlying representation of the group of words forming the paradigm, and the target MSD tag as a two-hot vector, and thus \textbf{does not need to make use of the surface form}. However, one problem with their model is that the system can't generalize to words from unseen paradigms, i.e. it can only complete paradigms which it has seen examples in the training data.

\newcite{silfverberg-hulden-2018-encoder} is in the same vein as \newcite{malouf2016} and \newcite{malouf2017} with more language acquisition consideration, but they use an \textbf{encoder-decoder model with soft attention} to solve the problem. More specifically, a 1-layer bidirectional LSTM is used to encode the given word form character by character, the source MSD features, and the target MSD features, and a 1-layer unidirectional LSTM with an attention mechanism is used to predict the target word form from the output of the encoder. Their model is more powerful for generalization, and can be used to complete paradigms which don't appear in the training data at all. In addition, \newcite{silfverberg-hulden-2018-encoder} conduct their experiments in a more restricted setting: they evaluated the model with one, two or three slots given in the paradigm respectively, and they also evaluated a situation resembling a language acquisition setting more closely, i.e. when only the most frequent word forms in each paradigm are given. For the cases where more than one form are given in a slot, they generate training pairs by pairing up given slots and thus reduce the paradigm cell filling problem to a string transduction problem; for the case where only one form is given in the paradigm, they first train an \textbf{LSTM language model} at the character level conditioned on MSD features with the training data, and then use the language model to guide a mechanism of character dropout in the inflection model to produce a kind of pseudo-stem. In this way, they can get 2 forms per table and convert the task to a (re)inflection task. 
When trained on the same amount of training data, the encoder-decoder model by \newcite{silfverberg-hulden-2018-encoder} achieved higher accuracy than their baseline model, i.e. the generator model by \newcite{malouf2017}, for most languages. For German nouns and Latin nouns in the case where only one form is given in each paradigm, the encoder-decoder model is slightly worse, which may be because the high degree of syncretism in the nominal declension of the two languages misleads the encoder-decoder model to copy too much, a problem that a model not relying on the surface word form, like the generator model, does not have.

Inspired by principal parts morphology, which states that a subset of forms in a paradigm, i.e. the principal parts, rather than a single form, provides enough information to determine the remaining forms of the paradigm \cite{finkel2007principal}, \newcite{cotterell2017graphical} induce topology and try to decode the entire paradigm jointly. Specifically, they construct a \textbf{paradigm tree} by using Edmond's algorithm \cite{edmonds1967optimum} to find the minimal directed spanning tree and the weight of the edges in the tree is the number of distinct edit paths between the characters in the pair of forms, which is calculated with the edit script procedure similar to \newcite{chrupala2008learning}. They use the \textbf{encode-decoder LSTM model with hard monotonic attention} \cite{aharoni2017morphological} to generate candidate target inflected forms and train the \textbf{graphical model} jointly to find the candidates for the missing slots that can best complete the paradigm. 
Their model was trained on more than 600 \textbf{complete paradigms}, an amount of data not available for most languages. 

\subsubsection{Low-resource scenario}

The second subtask of CoNLL-SIGMORPHON 2017 shared task \cite{cotterell-etal-2017-conll} is about paradigm completion, though only two teams submitted systems and results for this subtask \cite{kann2017sharedtask,silfverberg-etal-2017-data}. For training data, complete paradigms are provided, and for test, partial paradigms are provided and participants are expected to fill in the missing forms. High, medium, and low data settings, characteristic of CoNLL-SIGMORPHON 2017 shared task, are designed for this subtask as well. For the high data setting, most languages have around 200 full paradigms as training, and the number of inflected word examples varies as the size of paradigms are different for different languages. One of the participation teams \cite{kann2017sharedtask} focused on solving this problem. They use the encoder-decoder model architecture, specifically a bidirectional GRU as the encoder and a unidirectional GRU with an attention mechanism for decoding. They address the problem essentially as a multi-source string transduction task, similar to \newcite{kann-etal-2017-neural} and \newcite{cotterell2017graphical}. To improve the performance of the model, especially in \textbf{low-resource settings}, they use preprocessing and data augmentation. They preprocess the data by determining whether the language is predominantly prefixing or suffixing, using a special symbol to replace out-of-vocabulary (OOV) symbols at test time and copy the input symbol back for final prediction, a mechanism \newcite{silfverberg-etal-2017-data} also use. 
For \textbf{data augmentation}, they create training pairs by pairing up given forms in a paradigm with each other rather than using only pairs with the lemma and another inflected form, and add mappings of words to themselves and indicate it with a special tag, a technique they name as ``autoencoding''. They also experimented with first identifying the prefix or suffix of a word and replacing the stem part with random strings, a technique similar to the data hallucination method developed by \newcite{silfverberg-etal-2017-data}. In addition, they made use of the principal parts idea that members of a paradigm have different dependence on each other, and try to select the best source form when multiple forms are given according to \textbf{edit trees} \cite{chrupala2008learning}.  Specifically, for each paradigm, they pair up given slots and build edit trees for each pair of source form and target form. In this way, each target MSD tag would have a set of source MSD tags with a set of edit trees associated with it. The smaller the set of edit trees a source MSD tag has, the more important the source MSD slot is consider to be for the target MSD tag. Another approach \newcite{kann2017sharedtask} propose for the multi-source setting is fine-tuning, i.e. train one model for each language at each setting on all complete training tables, fine-tune the model on given slots in the partial tables at test time, and use the fine-tuned model to predict missing forms for the partial table. \newcite{silfverberg-etal-2017-data} is the other team to participate in this subtask, but it is not their focus. They also use the encoder-decoder model with an attention mechanism, but their encoder is a bidirectional LSTM and decoder is a unidirectional LSTM. A special symbol for OOV symbols at test time and the data hallucination method of replacing the stem with random strings is also used. Rather than using a majority vote to choose the best prediction from the output of multiple models, they use \textbf{weighted voting} with weights tuned by Gibbs sampling.

\newcite{kann-schutze-2018-neural} is also about paradigm completion in the \textbf{low-resource setting}, where being low-resource is defined as only a small number of \textbf{full paradigms} are available for training, same as CoNLL-SIGMORPHON 2017 shared task 2. Partial paradigms are provided at test time. Their model is based on the \textbf{encoder-decoder model with an attention mechanism} \cite{kann2016single}, improved by two methods which can be combined or applied separately: paradigm transduction and source selection. They found that in extremely low-resource settings where only 10 paradigms are available for training, the best performance was achieved by combining the method for source selection and the baseline of CoNLL-SIGMORPHON 2017 shared task, a non-neural model which extracts affixes based on character alignment; for cases with slightly more training data, like when 50 or 200 paradigms are available for training, combining the encoder-decoder model with both paradigm transduction and source selection achieves the best performance. For the paradigm transduction method, they first train the model on the training data as usual, and at test time, in addition to pairing up given forms, they also generate pairs where the given form with a special tag is mapped to itself, the same approach as autoencoding in \newcite{kann-schutze-2017-unlabeled}, and continue training the model with parameters initialized on the training data. For selecting the source forms, the same approach is used in \newcite{kann2017sharedtask} as explained above.

\subsubsection{Cross-lingual transfer}

The paradigm completion task in \newcite{kann2017one-shot} is a little different from the task in the papers mentioned above. In \newcite{kann2017one-shot}, the model generates all forms of the paradigm for a given lemma. They investigated the \textbf{cross-lingual transfer effect} for this task, i.e. to use training data for high-resource languages to help training models for low-resource language, where high-resource and low-resource are defined by the number of word inflection samples available for training. Their model is an \textbf{encoder-decoder} model, which is trained by appending a language tag to the input string. To be specific, the encoder is a bidirectional GRU and takes as input the concatenation of character embeddings for the lemma, MSD feature embeddings, and the representation for the language tag; the decoder is a unidirectional GRU with an attention mechanism. It's worth pointing out that MSD tags are broken down into features whose embedding representations are fed into the encoder. Their experiments on 21 languages from 4 language families indicate that cross-lingual transferring effect happens between languages that are closely related like Spanish and Portuguese as well as between languages that are very different, even with different writing systems, like Arabic and Spanish, though the more related the two languages are, the better morphological knowledge can be transferred. They also found that training with data from multiple languages can be helpful, though the performance improvements may not be obvious. For a tag which has been seen only once or hasn't appeared in the training data at all, if the high-resource language is very closely related to the low-resource language, the labeled data for the high-resource language can still help improve the prediction of the low-resource language. Though they have an experiment to separate the effect of true transfer from other effects by using a random cipher, it's not a strong support for cross-lingual transfer effect because the ciphered data also improve the prediction. 

\newcite{jin2017exploring} use the same model for the same task as \newcite{kann2017one-shot} and investigate the question of how and what knowledge gets transferred across languages. They conducted experiments on a subset of the data used in \newcite{kann2017one-shot} by focusing on the transferring effect of letters and tags. They found that knowledge of character sequences and MSDs can be transferred and the improvement by using training data from a high-resource language, even a not closely related language, is larger than a pure regularization effect. For both \newcite{kann2017one-shot} and \newcite{jin2017exploring}, the paradigm completion problem is treated the same as the task of \textbf{morphological inflection}.

\subsection{Segmentation with neural models}
\label{subsec:segment}

Most segmentation work to be discussed in this part involves segmenting parts of not only inflection, but also derivation and composition, which is the most common case for morphological segmentation, and derivation and composition tend to have a priority over inflection in labeled data for such tasks. 
For example, in the dataset published by \newcite{cotterell2016joint}, \textit{touching} is not segmented to \textit{touch ing}, rather it is taken as consisting of one morpheme. In other words, \textit{touching} is treated as an adjective, rather than the present participle form of \textit{touch}. The only exception is \newcite{moeller-hulden-2018-automatic}'s work on automatic glossing for language documentation, where they not only segment the affixes and the stem, but also label the segments with morphosyntactic descriptions and English glosses, a combination of segmentation and labeling.

\begin{figure}
    \centering
    \includegraphics[width=0.5\textwidth]{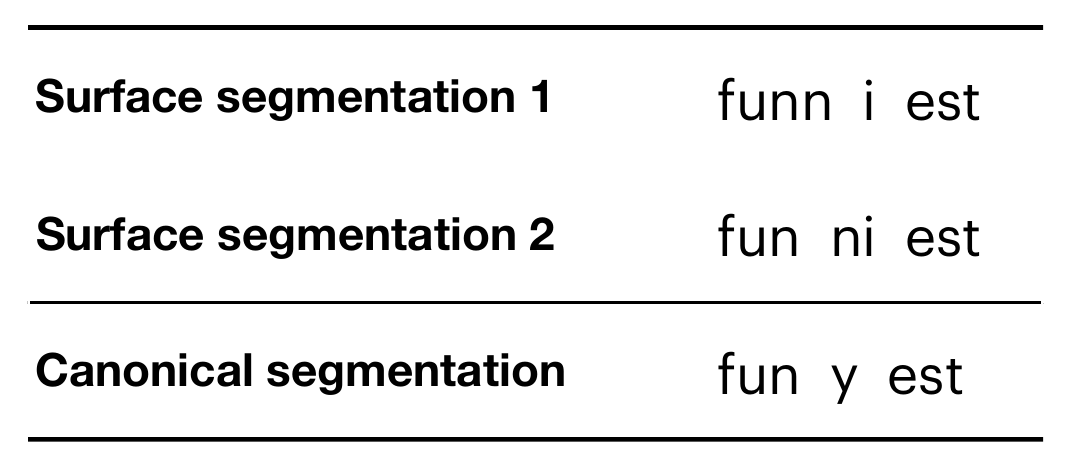}
    \caption{Two possible surface segmentations and the canonical segmentation of the English word \textit{funniest}. This example was created with reference to the example in \protect\newcite{cotterell2016joint}.}
    \label{fig:seg-example}
\end{figure}

\newcite{cotterell2016joint} distinguish surface segmentation and canonical segmentation. For surface segmentation, the words are split into multiple segments, and the concatenation of the segments are exactly the original word. In contrast, the segments are converted to canonical forms in canonical segmentation. Some ambiguity about segmentation boundary in surface segmentation can be avoided in canonical segmentation. However, the concatenation of the segments from canonical segmentation may be different from the original word and the canonical form for each morpheme needs to be identified. Identifying the canonical form for each morpheme is a task involving allomorphy learning discussed in section \ref{subsubsec:allomorph} on page \pageref{subsubsec:allomorph}. Figure \ref{fig:seg-example} provides two possible ways for doing surface segmentation for the English word \textit{funniest} and the canonical segmentation of the word.

\subsubsection{Surface morphological segmentation}

\newcite{wang2016morphological} is about \textbf{surface morphological segmentation}. They treat the task as a \textbf{sequence labeling} or structured classification problem, i.e. for each character in the word, to classify it as B (beginning of a multi-character morpheme), M (middle of a multi-character morpheme), E (end of a multi-character morpheme), or S (a single-character morpheme), a one-to-one learning scenario. They proposed three \textbf{window-based LSTM} architectures for this task with different objective functions: a simple window LSTM model that makes individual predictions for characters, a multi-window LSTM model which also captures the label transactions, and a bidirectional multi-window LSTM model which has a backward pass in addition to the forward pass added to the multi-window LSTM model. They conducted experiments on the Hebrew and Arabic data from the dataset by \newcite{snyder2008cross}, and designed settings of different training data sizes by first ordering the data based on the frequency of the word and using the 25\%, 50\%, 70\% or 100\% most frequent of the training data for experiments. They compared the performance of their models with regular LSTM models and non-neural models which require heavy linguistic feature engineering, and their model, especially 10-model ensembling of the bidirectional multi-window LSTM model outperforms the regular LSTM model by a large margin, and it also achieves comparable results to non-neural models, by inferring linguistic knowledge from data with the window LSTM neural models. 

\newcite{kann2018fortification} focus on the morphological segmentation task for polysynthetic languages with only a small amount of labeled data. They improve the character-based \textbf{seq2seq model with an attention mechanism} by leveraging \textbf{unlabeled data} or random strings. They utilize the additional data by applying a method for multi-task training which \textbf{jointly} maximize the log-likelihood of an auxiliary autoencoding task \cite{kann-schutze-2017-unlabeled} and the segmentation task, or by a \textbf{data augmentation} method which ``treat the amount of additional training examples as a hyperparameter ... [to] optimize on the development set separately for each language'' (p51). 
They also investigate the \textbf{cross-lingual transfer} effect by training one model for all related languages. In addition, they publish the dataset for four minimal-resource polysynthetic languages spoken in Mexico which they used for experiments. In general, their neural models did not achieve significant improvements over the CRF baseline they compared to.

Another related work aiming specifically at \textbf{low-resource languages} is \newcite{moeller-hulden-2018-automatic} who place the morphological segmentation and labeling task in the context of language documentation, where usually only a very limited amount of  data are available. Similar to \newcite{wang2016morphological}, for the CRF models they experimented with, they treat the task as a sequence-labeling problem, though their labels are much more details because their Beginning-Inside-Outside (BIO) tags are specified by morphological tags (as for fine-grained name-entity recognition where the model not only has to predict the scope of the named entity but also to specify what type of named entity it is) and they did not use a separate symbol for single-character morphemes. For the segmentation and labeling pipeline model, they still treat the segmentation task as a sequence labeling problem by using a CRF to predict the BIO tags, and then use a multi-class SVM to classify the morphemes as to their tags. In addition, they experimented with using an attention-enhanced encoder-decoder model with LSTM architectures for recurrence, which reads in the input word character by character and output the labels. This neural model did not surpass the CRF model with their designed features. The results of \newcite{kann2018fortification} and \newcite{moeller-hulden-2018-automatic} indicate that for conditions where we have extremely limited amount of labeled data, non-neural models with linguistic feature engineering still have an advantage, a finding also supported by \newcite{wolf2018structured} for morphological inflection within context.

\newcite{cotterell-schutze-2018-joint} incorporate semantics into segmentation. Their model does not have an overall neural network architecture. Instead, it consists of three factors: a transduction factor, which is a weighted-finite-state machine that computes probabilistic edit distance by looking at additional input and output context; a segmentation factor, which is an unnormalized first-order semi-CRF; and a composition factor, for which they experimented with addition and RNN respectively to combine morpheme embeddings to approximate the vector of the entire word. They found that \textbf{jointly} modeling semantic coherence and segmentation improves the segmentation performance, and that RNN is better than addition for vector composition.

\subsubsection{Canonical morphological segmentation}

\newcite{cotterell2016joint} introduce the canonical morphological segmentation task to convert a word to a sequence of segments in the ``canonical'' form. 
Their model is a joint model for transduction and segmentation with a finite-state transduction factor, 
a semi-Markov segmentation factor, 
and a sampling method to approximate the gradient for learning. An external dictionary is used in their model to check the validity of the segmentation. They evaluated their model on English, German and Indonesian and achieved state-of-the-art performance on the datasets they used and published for canonical morphological segmentation.

\newcite{kann2016neural} and \newcite{ruzsics2017neural} are both about \textbf{canonical morphological segmentation} with \textbf{attention-enhanced encoder-decoder models}, where the morphological segmentation task is treated as a \textbf{string transduction} problem, the same as \newcite{moeller-hulden-2018-automatic} did in their neural model experiment. \newcite{kann2016neural} add a neural reranker to the encoder-decoder model. The reranker uses external dictionaries and rescores the segmentation predictions with a feedforward network by incorporating explicit representations for entire segments with morpheme embeddings and thus making use of lexical information in addition to character information. They evaluated their model on the same dataset as \newcite{cotterell2016joint}, and achieved new state-of-the-art performance. \newcite{ruzsics2017neural} don't need the extra monolingual data employed by \newcite{cotterell2016joint} and \newcite{kann2016neural} and improve \newcite{kann2016neural}'s result on the same data for Indonesian and German. They achieve this by integrating a \textbf{language model} over the canonical segmentation into the decoder. To be specific, they train the standard encoder-decoder model with soft attention and the language model over morphemes individually on the training data. The encoder-decoder model predictions and the morpheme language model scores are combined to find the best segmentation with a beam search algorithm.


The use of external dictionaries in \newcite{cotterell2016joint} and \newcite{kann2016neural}, and the use of language models in \newcite{ruzsics2017neural} are efforts to utilize lexical meaning for morphological segmentation. All the work above did not make use of context information, though context is an important source of information for disambiguation when there are multiple ways to segment a word.

\begin{figure}
\centering
\includegraphics[width=.9\textwidth]{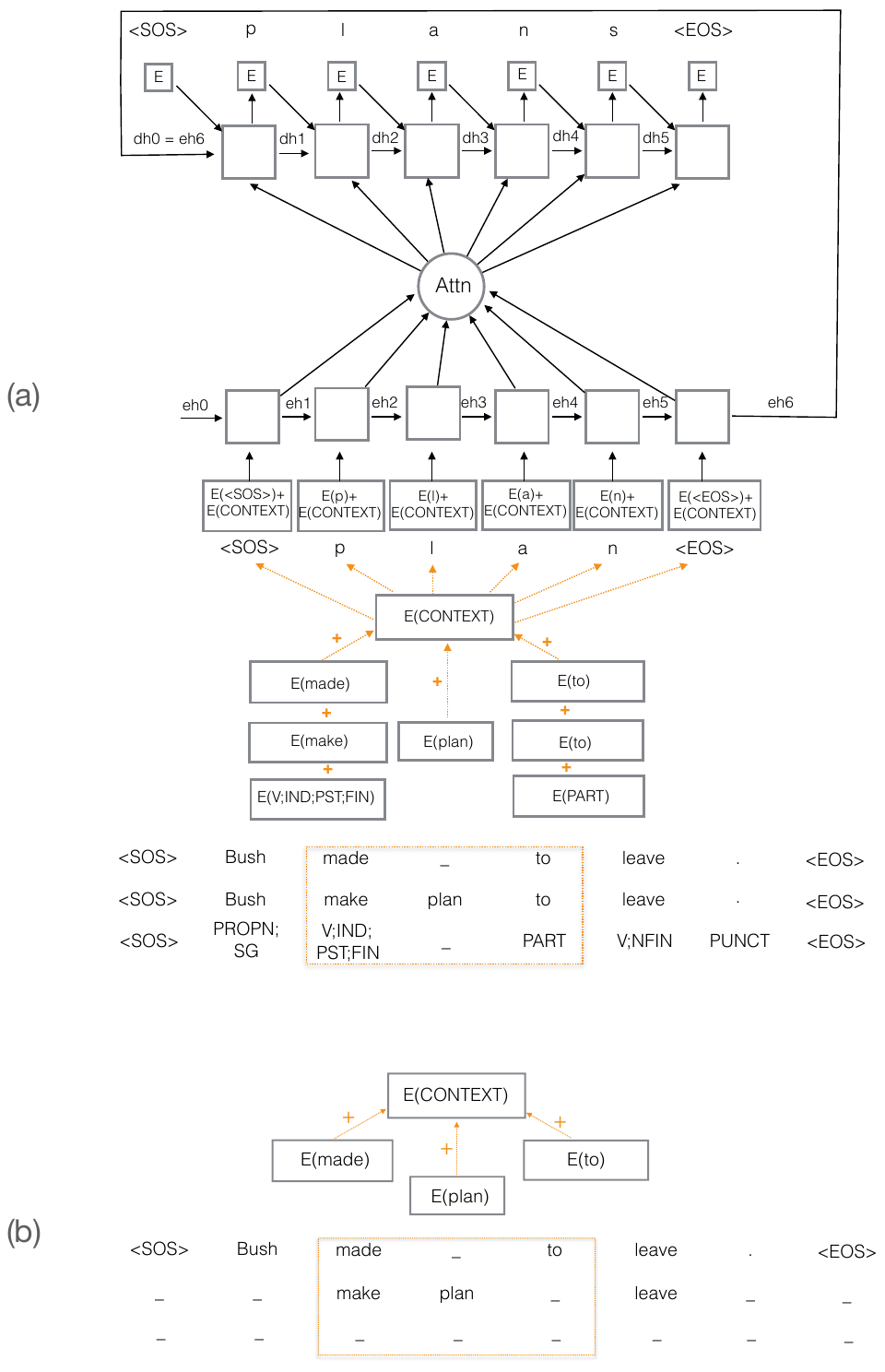}
\caption{Illustration of the seq2seq model applied to inflection in context. (a) illustrates the whole model architectures with labeled context (i.e. Track 1) (b) illustrates the contextual embedding for unlabeled context (i.e. Track 2). The figures are adopted from \protect\newcite{liu-etal-2018-morphological}.}
\label{fig:inflection-in-context-model}
\end{figure}

\section{Token-based computational morphology}
\label{sec:token-based}

After the success of neural network approaches with type-based morphological processing, more and more work starts to explore the performance of neural models to extract morphosyntactic information from sentential context. We see more applications of neural network approaches to token-based computational morphology in the past several years, especially morphological (re)inflection, lemmatization and morphological tagging in context. 

\subsection{Neural models and morphological (re)inflection in context}
\label{subsec:inflection_in_context}


The second subtask of CoNLL-SIGMORPHON 2018 shared task \cite{cotterell2018conll} is about morphological inflection in context, i.e. given a lemma and a sentential context it occurs in, including inflected forms of contextual words either with or without the corresponding lemma form and MSD tag, predict the correct inflected form for the lemma satisfying the context. The bottom three lines in Figures \ref{fig:inflection-in-context-model} (a) and (b) provide examples for the two versions of the task. This shared task also has three training data conditions: low, medium, and high. There were 6 submissions, all of which use neural models. The neural baseline system, illustrated in Figure \ref{fig:inflection-in-context-model}, has an \textbf{encoder-decoder with a soft attention mechanism architecture}, where the prediction of the inflected word form is conditioned on sentential context, different from type-based morphological inflection whose decoding is conditioned on the explicitly specified MSD tags. The context embedding is defined as a concatenation of embeddings of the current lemma at both word level and character level, the left and right context word form at the word level. When the lemma forms and the MSD tags for the left and right context words are given, They are also concatenated into the context embedding. Most submissions are developed on top of this baseline system, and define the context either as the baseline or expand the context by including more words before and after. \newcite{kementchedjhieva-etal-2018-copenhagen}'s system achieved the best overall accuracy of 49.87\%, and specifically it won this task in the high data setting for both tracks and the medium data setting for track 1. This system is an augmentation of the baseline system in three aspects: (1) a wider context -- an LSTM is used to encode the entire available context, (2) \textbf{multi-task learning} with the auxiliary task of MSD prediction for track 1, i.e. when the MSD tags for the contextual words are given, (3) first training models in a \textbf{multilingual} fashion with random groupings of two to three languages, followed by monolingual fine-tuning for each specific language. An ensemble of the five best models is used to make the final prediction by majority vote. \newcite{makarov2018conll} apply the system they propose in \newcite{makarov2018imitation}, which is a \textbf{neural transition-based transducer with a designated copy action} trained in a fully end-to-end fashion with the \textbf{immitation learning} framework \cite{daume2009search,ross2011reduction,chang2015learning}. 
This system won the task in the medium data setting for track 2 and the low data setting for both tracks. It's worth pointing out that even the overall best performing system only achieved an average accuracy of lower than 50\%, indicating that the problem is difficult and there is a large space for improvement.

\subsubsection{Low-resource scenario}

In order to utilize raw token-level data to improve morphological inflection in context in the \textbf{low-resource scenario}, \newcite{wolf2018structured} propose a \textbf{generative directed graphical model}, which is a full-fledged \textbf{language model} that can use both \textbf{labeled and unlabeled data}. The model consists of an LSTM language model for MSD prediction, a lemma generator defined as a conditional LSTM language model over characters, and a morphological inflector as a seq2seq model with a soft attention mechanism. When the model is trained with the raw text, unobserved lemmata and MSDs can be approximated by posterior inference over latent variables based on the wake-sleep algorithm \cite{hinton1995wake}. They found that agglutinative languages tend to benefit more from the use of raw text. One of their baselines is a probabilistic finite-state transducer system \cite{cotterell2017conll}. They found that the FST system is competitive with neural models when the training data is limited and that neural models can learn non-concatenative morphology better than the FST system. Though they claim to be the second attempt to perform semi-supervised learning for a \textbf{neural inflection generator}, preceded by \newcite{zhou2017multi}, it remains unclear why they didn't count other work trying to make use of unlabeled data as semi-supervised. Perhaps, they were talking about the model structure. Their model and \newcite{zhou2017multi}'s model both highlights a special design in the model architecture for leveraging unlabeled data, which is missing in models used in other work on neural morphological inflection.

\subsubsection{Joint learning}

\newcite{vylomova-etal-2019-contextualization} tackle CoNLL-SIGMORPHON 2018 shared task 2 with morphological tags given for contextual words. They propose \textbf{a structured neural model}, 
which explicitly reconstructs MSDs and then uses the predicted MSDs and lemmata to predict the corresponding inflected forms. Their tagger is a neural conditional random field and the morphological inflector is a neural encoder-decoder model with hard attention \cite{aharoni2017morphological}. They compared their system with a system that directly predicts the inflected forms without relying on any morphological annotation, as well as the shared task baseline system \cite{cotterell2018conll} and the monolingual version of the shared task winning system \cite{kementchedjhieva-etal-2018-copenhagen}. Their lemmatization results outperform other models for most languages. They found that when lemmatization is conditioned on the ground-truth MSD tags, the accuracy can be much higher. These results indicate that \textbf{jointly} predicting morphological tags is helpful for morphological inflection in context. This supports incorporating linguistically motivated information into neural models.

\subsection{Neural models and context-sensitive lemmatization}
\label{subsec:context_and_lemmatization}

\begin{table}
    \centering
    \begin{adjustbox}{width=.85\textwidth}
    \begin{tabular}{llcccccccc}
        \toprule
        \textbf{Input sentence} & He & told & the & children & an & encouraging & story & . \\
        \textbf{Lemmatization} & he & tell & the & child & a & encouraging & story & . \\
        \midrule
        \textbf{Input sentence} & He & is & encouraging & the & children & with & a & story & . \\
        \textbf{Lemmatization} & he & be & encourage & the & child & with & a & story & . \\
        \bottomrule
    \end{tabular}
    \end{adjustbox}
    \caption{Examples for lemmatization in context. The context is needed to determine whether \textit{encouraging} should be lemmatized as \textit{encourage} or not.}
    \label{tab:contextual_lemmatization}
\end{table}

Table \ref{tab:contextual_lemmatization} provides examples of context-sensitive lemmatization where the task is to convert the words in the given sentence to the corresponding lemma forms satisfying the context. Neural approaches to lemmatization in context have been treating the task either as \textbf{a classification problem of selecting and applying an edit tree that produces the lemma from the inflected form} or as \textbf{a string transduction problem of mapping the input character string of the inflected word forms to the lemma string}. 

\newcite{chakrabarty2017context} treat the task as an \textbf{edit tree classification} problem with two successive bidirectional RNNs. The first bidirectional RNN builds the syntactic embedding, i.e. to generate the composite vectors for all words in the training set, and the second bidirectional RNN learns the representation of the local context and is trained to predict the most likely lemma corresponding to the applicable edit trees. In other words, the second bidirectional RNN predicts the lemma for the word in context by identifying the correct edit tree as a classification problem. They experimented with both LSTM and GRU for the recurrent neural network architecture, and found that they performed equally well. The system outperformed Lemming \cite{muller2015joint} and Morfette \cite{chrupala2008learning}, two previous state-of-the-art models, both of which are non-neural models learning to do lemmatization and morphological tagging jointly, on 9 languages except Bengali. The annotated Bengali dataset is another contribution by \newcite{chakrabarty2017context}.

\begin{table}
\centering{
\begin{adjustbox}{width=0.7\textwidth}
\begin{tabular}{lll|lll}
\toprule 
run & run & V;NFIN & speak & speak & V;NFIN \\
run & ran & V;PST & speak & spoke & V;PST \\
run & running & V;V.PTCP;PRS & speak & speaking & V;V.PTCP;PRS \\
run & runs & V;3;SG;PRS & speak & speaks & V;3;SG;PRS \\
run & run & V;V.PTCP;PST & speak & spoken & V;V.PTCP;PST \\
\bottomrule
\end{tabular}
\end{adjustbox}
\caption{UniMorph inflection table examples with English verbs \textit{run} and \textit{speak}. For each verb, the first column is the lemma, the second column is the inflected form and the third column is the corresponding MSD tag.}
\label{tab:unimorph-example}
}
\end{table}

\subsubsection{Low-resource scenario}

The model \textit{Lematus} \cite{bergmanis2018context} is a lemmatizer based on the standard \textbf{encoder-decoder with soft attention mechanism} \cite{kann2016med}, and incorporates sentential context. The system was developed especially with the consideration to improve the lemmatization of \textbf{ambiguous words} and \textbf{unseen words}. They also conducted experiments with different amount of training data, and experimented with different ways to represent context, including word-level context representations that use all words in the left and the right of the sentence, character-level context representations that use 0, 5, 10, 15, 20, or 25 characters in the left and right of the sentence, and sub-word unit context representations by byte pair encoding. For the 20 languages from UD v2.0 they tested on, \textit{Lematus} outperformed Lemming \cite{muller2015joint}, Morfette \cite{chrupala2008learning}, \newcite{chakrabarty2017context} and a lookup-based baseline which predicts the most frequent lemma form for observed data and the word itself for unseen words, in full-data setting and comparable results in limited data setting (i.e. with 10k training examples). They found that including context can improve the performance of the system significantly in both full data setting and limited data setting, and that context can help more with the lemmatization of ambiguous words than unseen words. Their cross-linguistic analysis indicates that the proportion of unseen words in a language is anti-correlated to the proportion of ambiguous words in that language. \newcite{bergmanis-goldwater-2019-training} attempt to improve the performance of \textit{Lematus} when the training data is limited by using UniMorph \cite{kirov2018} inflection tables (see Table \ref{tab:unimorph-example} for example) and Wikipedia text to create additional training data. Their results indicate that this \textbf{data augmentation} method is useful in low- and medium-resource settings.

\subsubsection{Joint learning}

Lemmatization and morphological tagging in context are mutually dependent and can provide information for each other to exclude unlikely choices in cases where there are multiple ways for lemmatization or tagging. It's very common to conduct lemmatization and morphological tagging together, which is usually referred to as morphological analysis where morphological analysis is used in the narrower sense. Table \ref{tab:contextual_lemmatization_tagging} adds the ground-truth MSD tags to each word in the examples in Table \ref{tab:contextual_lemmatization}. The \textit{encouraging} in the first example is an adjective whose lemma form should be \textit{encouraging}, while the \textit{encouraging} in the second example is a verb in the present participle form whose lemma form should be \textit{encourage}. \newcite{muller2015joint} provides empirical evidence that joint modeling of lemmatization and morphological tagging is mutually beneficial. There are also neural models developed for \textbf{joint} learning of lemmatization and morphological tagging, e.g. \newcite{kondratyuk-etal-2018-lemmatag}, \newcite{malaviya-etal-2019-simple}, \newcite{mccarthy-etal-2019-sigmorphon}, which provide further support for the effectiveness of learning context-sensitive lemmatization and morphological tagging jointly.

\begin{table}
    \centering
    \begin{adjustbox}{width=\textwidth}
    \begin{tabular}{lccccccccc}
        \toprule
        \textbf{Input sentence} & He & told & the & children & an & encouraging & story & . \\
        \textbf{Lemmatization} & he & tell & the & child & a & encouraging & story & . \\
        \textbf{MSD tagging} & PRON;NOM;SG & V;PST & DET;DEF & N;PL & DET;IND & ADJ & N;SG & PUNCT \\
        \midrule
        \textbf{Input sentence} & He & is & encouraging & the & children & with & a & story & . \\
        \textbf{Lemmatization} & he & be & encourage & the & child & with & a & story & . \\
        \textbf{MSD tagging} & PRON;NOM;SG & AUX;SG;3;PRS & V;PRS;PCTP & DET;DEF & N;PL & ADP & DET;IND & N;SG & PUNCT \\
        \bottomrule
    \end{tabular}
    \end{adjustbox}
    \caption{Examples for lemmatization and morphological tagging in context.}
    \label{tab:contextual_lemmatization_tagging}
\end{table}

\textit{LemmaTag} \cite{kondratyuk-etal-2018-lemmatag} is a bidirectional RNN architecture with character-level and word-level embeddings that jointly generates MSD tags and lemmata for sentences. Specifically, the model consists of three parts: (1) The encoder with the structure of \newcite{chakrabarty2017context}. A bidirectional RNN gets the character-level representation of each word \cite{heigold-etal-2017-extensive} and the embedding for each character are concatenated to form the character-level embedding for the word. The character-level embedding is summed up with the word-level embedding obtained from a word-level embedding layer to form the final word embedding for the word. The final word embeddings are the input into two layers of bidirectional RNN with residual connections to get the context representation. The encoder is shared by the tagger decoder and the lemmatizer decoder. (2) The tagger applies a fully-connected layer to the representations obtained with the encoder to generate the tags. (3) The lemmatizer applies an RNN decoder with soft attention to character encodings, summed embeddings for the current word, predicted tag features and context representations to generate the lemma form character by character. The final loss function is defined as the weighted sum of the losses of the tagger and the lemmatizer. This system achieved state-of-the-art performance on tagging and lemmatization in Czech, German and Arabic, and comparable performance to the state of the art for English. They point out that jointly learning lemmatization and morphological tagging by sharing encoder parameters and providing the predicted tags as input to the lemmatizer improves the performance of both lemmatization and tagging, especially for morphologically rich languages. However, in languages with less complex morphology like English and German, it may hurt the performance of the tagger to share the encoder parameters.

The joint neural model for lemmatization and morphological tagging developed by \newcite{malaviya-etal-2019-simple} has a similar architecture with \newcite{vylomova-etal-2019-contextualization}, i.e. a \textbf{structured neural model} consisting of a morphological tagger part and a lemmatizer part. The morphological tagger is the SPECIFIC model from \newcite{cotterell-heigold-2017-cross} reimplemented by \newcite{malaviya-etal-2018-neural}. The lemmatizer is a seq2seq model with hard attention mechanism. Greedy decoding and crunching \cite{may-knight-2006-better} were experimented with to get the tagging result used for lemmatization. The crunching scheme is a heuristic to approximate true joint learning by first predicting the k-best tags with the morphological tagger and then incorporating the predicted tags into the lemma generation process. The crunching schema achieved slightly higher average accuracy. They train the models with jackknifing \cite{agic-schluter-2017-train} to avoid exposure bias in order to improve the lemmatization result. The evaluation focuses on the lemmatization results. Their lemmatization results outperform \textit{Lematus} \cite{bergmanis2018context}, UDPipe \cite{straka-strakova-2017-tokenizing}, Lemming \cite{muller2015joint} and Morfette \cite{chrupala2008learning}. Similar to the finding of \newcite{kondratyuk-etal-2018-lemmatag}, their error analysis indicates that jointly learning morphological tagging and lemmatization is especially helpful for languages with more complex morphology. They also found that when the training data is limited for lemmatization, joint learning of morphological tagging and lemmatization is helpful. Similar to \newcite{vylomova-etal-2019-contextualization}, \newcite{malaviya-etal-2019-simple}'s experiments show that when the ground-truth MSD tags are given to the lemmatizer, the lemmatization accuracy can be much higher. 

The second subtask of SIGMORPHON 2019 shared task \cite{mccarthy-etal-2019-sigmorphon} is on morphological tagging and lemmatization in context. The non-neural baseline for this task is Lemming \cite{muller2015joint} and the neural baseline is \newcite{malaviya-etal-2019-simple}. 16 systems were submitted for this subtask, all of which use neural network models. The most successful systems are implementations of variations of multi-headed attention, multi-level encoding and multiple decoder, and use BERT \cite{devlin-etal-2019-bert} pretrained embeddings for contextual representations. \newcite{kondratyuk-2019-cross}'s system achieved the highest average accuracy and F1 score for morphological tagging and the second average accuracy in lemmatization. It consists of a shared BERT encoder and predicts lemmata and MSD tags jointly. To be specific, BERT multilingual cased tokenizer is first applied to each token in the input sentence, which is then encoded with pretrained multilingual cased BERT base model. Two separate layer attention instances similar to ELMo \cite{peters-etal-2018-deep} as defined in UDify \cite{kondratyuk-straka-2019-75}, generate embeddings specific to each task. A bidirectional LSTM with residual connection is applied to the sequence of characters for each word to encode the morphological information, and two successive layers of word-level bidirectional residual LSTMs are leveraged to compute over the entire task layer attetion for each task. For lemmatization decoding, it is treated as a \textbf{sequence classification} task to find the \textbf{edit actions} to transform an inflected word to its lemma and a feedforward layer is applied to the lemmatizer LSTM final layer to get the result. For morphological tagging, a feedforward layer is applied to jointly predict the factored and unfactored MSDs. Factored MSDs treat the MSD tags at the feature tag level, while unfactored MSDs treat the whole chunk of a MSD tag as one tag. For example, for a MSD tag \textit{N;PL}, there are two factored tags \textit{N} and \textit{PL} while it is one unfactored tag. The factored tags are used to improve training and the tagging prediction are the output of the unfactored tags. Their results support the \textbf{cross-lingual transfer} effect for boosting lemmatization and morphological tagging because they achieve better performance by fine-tuning multilingual BERT on all available treebanks for all languages over fine-tuning monolingually. In addition, they found that the additional character-level and word-level LSTM layers for embedding can improve the performance even further. 

The winning system of the lemmatization task in SIGMORPHON 2019 shared task 2 is \newcite{straka-etal-2019-udpipe}, which ranks the second for morphological tagging performance. This system is an improvement of UDPipe 2.0 \cite{straka-2018-udpipe} which uses a single joint model for morphological tagging and lemmatization with bidirectional LSTM models. 
UDpipe also treats lemmatization as a \textbf{classification} problem by making the model to predict generation rules to convert the inflected word forms to their corresponding lemmata. The improvements are from 3 aspects. First, similar to \newcite{kondratyuk-2019-cross}, they use pretrained BERT embedding as additional inputs to the network. Secondly, also similar to \newcite{kondratyuk-2019-cross}, they regularize the model by predicting factored tags though they take the tagging prediction as the unfactored tags. Thirdly, rather than concatenating all languages like \newcite{kondratyuk-2019-cross}, they merge treebanks for the same language. The success of \newcite{kondratyuk-2019-cross} and \newcite{straka-etal-2019-udpipe} indicates the effectiveness of predicting \textbf{edit operations} rather than character by character for lemmatization, a finding also supported by \newcite{yildiz-tantug-2019-morpheus} who submitted the third ranked lemmatization system for this task. 

\newcite{shadikhodjaev-lee-2019-cbnu} is another team participating in SIGMORPHON 2019 shared task 2. They developed a \textbf{transformer-based seq2seq model} for lemmatization and a bidirectional LSTM model with attention for morphological tagging. They produce the lemma for each word in the sentence, and then use the predicted lemma and the given word form together to predict the MSDs. This way of using lemmatization output to improve tagging as a pipeline is not common, and the more common way is the other around. Their system ranked fifth in lemmatization and sixth in tagging.

Another work to jointly learn lemmatization and morphological tagging is \newcite{akyurek2019morphological}. This system focuses on morphological tagging and the tagging results outperform \newcite{cotterell-heigold-2017-cross} and \newcite{malaviya-etal-2018-neural}. It's interesting to point out that they found that making the decoder to jointly predict the lemma form and the factored MSDs produces generally worse results than decoding only the MSDs, especially in the low-resource setting and for morphologically more complex languages. They also found that predicting factored MSDs is advantegous over predicting unfactored MSDs when the training data is limited and the language has rich inflections, and that using related high-resource language data to boost \textbf{low-resource language} training is helpful. The structure of their system is as follows: They use a unidirectional LSTM at the character level to get the word embedding for each word; a forward LSTM at the word level to read in words to the left of the current word and a word-level backward LSTM to read in words to the right of the current word, and take the concatenation of the forward and backward LSTMs as the context representations; another unidirectional LSTM to encode the predicted morphological tagging of the previous two words. For the decoder part, they use an LSTM to generate the lemma character by character followed by the MSD features. In their model, they did not make use of the lemma information other than in the decoding process, which explains why the model generally performs better when it only predicts the MSDs. Another contribution of the work is that they published a new Turkish morphology dataset with a high (96\%) inter-annotator agreement. 

Closely related to joint learning of lemmatization and morphological tagging is \textbf{morphological disambiguation}, as shown with the examples in Table \ref{tab:contextual_lemmatization_tagging} where we need to disambiguate \textit{encouraging} in order to get the correct lemma form and MSD tags. \newcite{yildiz2016disambiguation} propose a neural model which can pick out the best lemma and MSD analysis for each word in sentential context, but they rely on existing morphological analyzers to first produce the possible lemmata and MSDs.

Another vein of lemmatization is the \textbf{lemmatization of non-standard languages}. \newcite{kestemont2016lemmatization} handles lemmatization of Middle Dutch with neural models. Specifically, they use a character-level CNN to learn token representations, and a neural encoder to extract character- and word-level features from a fixed-length token window of 2 words to the left and 2 words to the right, which are then used to predict the target lemma from a closed set of true lemmata. \newcite{manjavacas-etal-2019-improving} take a \textbf{joint} learning approach to improve the lemmatization of non-standard historical languages, where the sentence encoder is trained jointly for lemmatization and \textbf{language modeling}. Specifically, they take lemmatization as a string transduction task with an encoder-decoder architecture and use a hierarchical sentence encoder similar to \newcite{kondratyuk-etal-2018-lemmatag} to enrich the encoder-decoder architecture with sentence context information. Their system does not require POS and MSD tagging. With the joint lemmatization and language modeling sentence encoder, their system can make use of both character-level and word-level features of each input token, a more general approach to use a predefined fixed-length window of neighbouring tokens, like in \newcite{kestemont2016lemmatization} and \newcite{bergmanis2018context}. They compared their system with a simple seq2seq model without sentence-level information, a model without joint language modeling loss and two state-of-the-art non-neural models using edit-tree induction, Morfette \cite{chrupala2008learning} and Lemming \cite{muller2015joint}. Their system achieved state of the art performance on historical language lemmatization and comparable performance on standard languages. They found that the language modeling loss is helpful for capturing morphological information, and edit-tree-based approaches can be very effective for highly synthetic languages while the encoder-decoder architecture are more effective for languages with higher ambiguity and token-lemma ratio. The eight corpora of historical language they use are medieval and early modern languages, including Middle Dutch, Middle Low German, Medieval French, Historical Slovene and Medieval Latin, and they published the datasets.

\subsection{Neural models and morphological tagging}
\label{subsec:model_and_tagging}

Morphological tagging is usually conducted on words in sentences, making this a sequence labeling problem,
where for each word in the sentence the tagger is expected to predict the corresponding MSD tag. This corresponds to the one-to-one learning scenario. The use of information from the context is critical for good tagging performance. 


The typical neural model for this task has three stages: (1) word representation part, (2) context encoding part, (3) tagging part. \newcite{labeau-etal-2015-non} use a convolutional layer to derive word representations from a sequence of character embeddings, and compare the effect of using such character-based embeddings with using only conventional fixed-vocabulary word embeddings. They experimented with the feedforward architecture and bidirectional LSTM architecture for the context encoding part. For the tagging part, they experimented with two approaches: a simple softmax prediction, i.e. to apply a softmax function to the output at each time step from the context encoding to predict the corresponding tag for that time step; and a structured prediction approach, where they use the Viterbi algorithm \cite{viterbi1967error} to infer the most likely MSD tag one feature after another. They evaluated their model with the German corpus TIGER Treebank \cite{brants2002tiger}. The best result was achieved by combining the character-based character embedding derived with a CNN layer and the conventional word embedding, using bidirectional LSTM to encode the context and applying the Viterbi algorithm to predict the tag sequence in a structured way. Their model outperformed the previous state-of-the-art model \cite{mueller-etal-2013-efficient} on the same dataset, which is a high-order CRF model with intensive feature engineering. \newcite{heigold2016neural} and \newcite{heigold-etal-2017-extensive} focus on evaluating different ways to get word representations on different languages, including feedforward, CNN, CNNHighway, LSTM and bidirectional LSTM. Their models use bidirectional LSTM to encode the context and a simple softmax layer applied to each time step to predict the tags. They achieved even better result. \newcite{heigold2016neural} found that when carefully tuned, the difference between the embedding models is minor, and \newcite{heigold-etal-2017-extensive} found that the LSTM-based approach is slightly better and more consistent than CNN-based approaches. 



\subsubsection{Cross-lingual transfer}

\newcite{cotterell-heigold-2017-cross} 
apply the model in \newcite{heigold-etal-2017-extensive} for \textbf{cross-lingual transfer} learning. They train an RNN tagger for a low-resource language jointly with a tagger for a related high-resource language and the models for low- and high-resource languages share character-level features. They experimented with three cross-lingual transfer learning scheme: language-universal softmax, language-specific softmax, and joint morphological tagging and language identification. In the language-universal softmax scheme, they add a symbol as language identifier before and after the word and use the new form as input to the bidirectional LSTM embedding layer. In the language-specific softmax scheme, the model is trained on the combined data of all the languages, but has an independent output layer for every language. For the joint learning setting, the model predicts the language and MSD tag together. They found that the shared character representations among related languages successfully enable knowledge transfer between these languages. Their model, especially with the joint morphological tagging and language identification scheme, outperforms the MARMOT tagger \cite{mueller-etal-2013-efficient}, and the tagger by \newcite{buys-botha-2016-cross}. 
\newcite{buys-botha-2016-cross} also investigate morphological tagging for low-resource languages by cross-lingual transfer. The model they propose is a neural discriminative model based on Wsabie \cite{weston2011wsabie}. Their approach is projection-based and the neural model learns to rank the set of tags allowed by the projection. The projection is conducted as a separate stage before the neural network modeling, and a separate morphological tagger is needed for the source-language text. In contrast, the approach of \newcite{cotterell-heigold-2017-cross} for cross-lingual transfer is end-to-end. Though compared to \newcite{cotterell-heigold-2017-cross} which still requires a small amount of data for the low-resource language with direct MSD annotation, \newcite{buys-botha-2016-cross} don't have this requirement, the parallel text between high-resource languages and low-resource languages can still be costly to get and the performance of the supervised morphological tagger for the source language is critical for obtaining high-quality constraints for the target language tagging.

\newcite{malaviya-etal-2018-neural} point out that the model of \newcite{cotterell-heigold-2017-cross} has the shortcoming of only being able to output tags that have appeared in the training data and that the tagsets should be shared between the languages for the knowledge transfer. They propose a factorial conditional random field \cite{sutton2007dynamic} combined with bidirectional LSTM method which improves in these two aspects. Their approach achieved better performance than \newcite{cotterell-heigold-2017-cross}. In addition, the graphical approach can model dependencies between tags for the whole unfactored MSDs and dependencies between unfactored MSDs in the sentence with better interpretability.

\newcite{kirov-etal-2017-rich} is also about \textbf{cross-lingual transfer} effect, and it is also projection-based and requires parallel data as \newcite{buys-botha-2016-cross}, but the research question is different: \newcite{kirov-etal-2017-rich} aim at verifying the linguistic claim about equal expressiveness of languages, i.e. whether it's true that what is conveyed by syntactic or contextual cues in languages with comparatively limited morphology is expressed by overt inflection or derivation in languages with rich morphology. They first project the complex morphological tags of Czech words directly onto the English words they align to in the Prague Czech-English Dependency Treebank (PCEDT) \cite{hajic-etal-2012-announcing}, and then develop a neural network tagging model with the feedforward architecture that takes input English features, including basic features derived directly from text, features derived from dependency trees and features from CFG parsing. The neural model attempts to classify the English word into the projected Czech tags. The high classification accuracy produced with the English features, gives support to the linguistic hypothesis of equal expressivity. 

\newcite{conforti-etal-2018-neural} is different from the tagging papers discussed above because they tag sequence of lemmata with part-of-speech and morphological features, rather than tag surface word forms. They presuppose the data has been lemmatized and use Lemming \cite{muller2015joint} to preprocess the data. Their tagging model is three layers of bidrectional GRU which takes the sequence of lemmata as input and predict the tags for each lemma with a softmax layer. They point out that their work can be helpful for machine translation.

\subsubsection{Joint learning}

In addition to \textbf{joint} training of morphological tagging and language identification \cite{cotterell-heigold-2017-cross}, it has been common to jointly model morphological tagging and lemmatization (specifically, lemmatization in context), which has been discussed in the previous subsection on context-sensitive lemmatization.

\section{Discussion}
\label{sec:discuss}

\subsection{Advantages of neural models}

Neural network models have been very successful in processing morphology, and have improved the performance of almost all tasks in computational morphology by a large margin over previous state-of-the-art non-neural models. This success was especially impressive in SIGMORPHON 2016 shared task \cite{cotterell2016sigmorphon} of morphological (re)inflection which received both neural and non-neural system submissions with all best-performed models being neural and the best neural model outperforming the best non-neural model by over 10\% in average accuracy. Neural models have become dominant in all computational morphology tasks. Though neural models are data-hungry and require a large amount of labeled data to achieve good performance, later research on improving neural network models in low-resource scenarios also turn out to be useful and neural models with augmentations can still produce results better than or comparable to non-neural models when the training data is limited. 

The success of deep learning methods is especially impressive with generation tasks in computational morphology. Before neural models became popular, work in computational morphology focused more on analysis rather than generation \cite{malouf2016,malouf2017}. Generation work before neural network includes \newcite{durrett2013supervised}. 

Non-concatenative morphology like in Semitic languages has been a problem non-neural models can't handle well. Neural models have been very successful in this aspect \cite{wolf2018structured}. In addition, neural models have been more flexible and creative, and thus can deal with unseen words or MSD tags as well as ambiguous words much more effectively \cite{bergmanis2018context,akyurek2019morphological}. 

In contrast to the much better performance is the much less feature engineering. Finite state-machine based on rules written by human experts, which is very time-consuming and expensive \cite{beemer-etal-2020-linguist}. Statistical machine learning approaches such as CRFs, SVMs usually require heavy feature engineering and heuristics inspired by language expert knowledge. Neural models usually don't need such feature engineering. In addition, they provide a way to replace traditional features in finite-state machine based approaches and statistical machine learning based approaches. For example, \newcite{rastogi2016weighting} infuse features encoded with neural models into weighted FSTs for morphological inflection; graphical models like \newcite{cotterell-heigold-2017-cross}, \newcite{wolf2018structured}, and \newcite{vylomova-etal-2019-contextualization} incorporate neural network approaches into statistical machine learning methods.

Another advantage of deep learning approaches is that they can incorporate contextual information and meaning (inferred from context, or using embedding information) more effectively. As shown in section \ref{sec:token-based}, neural models for token-based tasks in computational morphology have achieved state-of-the-art performance, and tasks like morphological (re)inflection and lemmatization in context were rarely tackled before neural network models became popular for computational morphology. 

Neural network models also make it possible to explore questions that are hard to explore before. For example, the cross-lingual transfer effect has been a phenomenon receiving wide research, thanks to neural network models. Can we use labeled data in high-resource languages to boost the model training for low-resource languages? How closely related should the high-resource language be in order to be helpful for the low-resource languages? These questions have been one of the focus of SIGMORPHON shared tasks \cite{mccarthy-etal-2019-sigmorphon} and received wide interest. Tagging tasks have seen improvements in models for low-resource languages when models are trained with related high-resource languages (e.g. \newcite{buys-botha-2016-cross}, \newcite{cotterell-heigold-2017-cross}, \newcite{akyurek2019morphological}, and \newcite{kondratyuk-2019-cross}), while high-resource languages are found to contribute to the (re)inflection models for low-resource languages in some work (e.g. \newcite{kann2017one-shot}, \newcite{jin2017exploring}, \newcite{anastasopoulos-neubig-2019-pushing}), and don't in some others (e.g. \newcite{bergmanis2017training}, \newcite{rama-coltekin-2018-tubingen}, \newcite{coltekin-2019-cross}, \newcite{hauer-etal-2019-cognate}, \newcite{madsack-weissgraeber-2019-ax}). Though we still lack an effective evaluation method to determine whether the improvement in performance is actually transfer of linguistic knowledge between languages or just regularization or some other reasons, it's still exciting that neural models add this flexibility in using labeled data.

The end-to-end learning process of neural network models contributes to multi-task learning and joint learning in computational morphology. In addition, it provides an alternative way to model the acquisition of morphology, with the learning curve and result of neural approaches being similar to human language acquisition findings \cite{rumelhart1986learning,kirov-cotterell-2018-recurrent}, though the adequacy of neural networks as a good cognitive model is still questionable \cite{corkery-etal-2019-yet}.

\subsection{Problems and future directions}


It usually requires a good amount of examples in order to train a good neural network model. \newcite{liu2021can} find that when measuring the amount of the training examples for morphological tasks, we should count the lemmata rather than inflected forms: a good representation of different lemmata is critical for the model to learn generalization over morphology.

In addition to being data-hungry, there are other problems with neural network approaches. One notorious problem that people have been trying to overcome, is the lack of interpretability. Current efforts to improve the interpretability in the learning of morphology is mainly to speculate what neural models have learned by error analysis of the output, ablation analysis of the model to see how much each part of the model contributes to the performance, plotting attention weights to see how much different parts of the input contribute at each step for generating the final prediction with the hope to find the correlation between the output and the parts attended to or plotting embedding representations with dimension reduction. Such interpretability alone can't contribute to better understanding of human language. More work needs to be done to interpret neural models as well as the understanding of human languages from the linguistics and cognitive science perspectives.

Related to model interpretability is the contribution of linguistic knowledge to computational processing of morphology. Though neural network approaches do not require high-level feature engineering, incorporating linguistically motivated knowledge has been found helpful for computational morphology tasks, e.g. \newcite{silfverberg-etal-2018-sound,vylomova-etal-2019-contextualization,malaviya-etal-2019-simple,liu-hulden-2020-leveraging}). It would be beneficial to develop tools facilitating linguistic studies as well if neural models can leverage and improve upon linguistic knowledge. This is a direction where more future work is necessary.

Morphology has an interface with syntax, which is reflected mainly in token-based computational morphology tasks, i.e. morphological processing of words in sentential context. This aspect has seen great progress with neural network approaches. Another interface of morphology with the language system is the interface between morphology and sounds, i.e. phonetics and phonology. Little work has been done with neural models as to allophonic and allomorphic analysis. This may be due to the lack of annotated data in this aspect, and labeling such data requires expert linguistic knowledge, especially on phonetics and phonology, making the labeling really expensive. With the end-to-end training fashion of neural models, perhaps some attempts can be conducted to use audio to assist the computational processing of this interface. In addition, though neural network models have been dealing with different writing systems well, even for learning cross-lingual information, in order to explore more about the interface between morphology and phonology, a phonemic representation of words rather than an orthographic one would be very helpful. Grapheme-to-phoneme (G2P) conversion systems (e.g. \newcite{ott2019fairseq}) can be used for this purpose.


Unsupervised learning was a large part of computational morphology before neural models \cite{goldsmith2017computational,hammarstrom2011unsupervised,ruokolainen-etal-2016-comparative}, which was primarily on segmentation. 
With neural network model, the work closely related to unsupervised word segmentation is on subword units. \newcite{bojanowski-etal-2017-enriching} use character n-grams to enrich word vectors, which turns out to be very effective, especially to deal with rare words. \newcite{sennrich-etal-2016-neural} apply the byte pair encoding (BPE) algorithm \cite{gage1994new} to get subword units, and show that segmentation with BPE improves over character n-gram models. BPE has been the commonly used segmentation method to get subword units for downstream processing with neural network models, though \newcite{liumt2021} found that for morphologically complex language, segmentation by syllables has some advantage in machine translation. 
However, such unsupervised segmentation are not intended to learn segment boundaries that matches morpheme boundaries. The effort and use of neural models to learn actual morpheme boundaries in an unsupervised way is quite limited. More work can be done as to unsupervised learning of morphology with neural models in the future, especially considering the current success of deep learning approaches with token-based morphology tasks, which indicates that neural models can infer morphosyntactic information well from context.  

There are other linguistic questions which can be explored more with neural models. One of such questions is about the equal expressiveness of languages, i.e. whether what is expressed in one language with morphology can be equally expressed in another language with syntactic or discourse techniques. \newcite{kirov-etal-2017-rich} is a paper on this where they first project Czech MSD tags to aligned parallel English words in the PCEDT corpus, and develop a neural model predicts the complex MSD tags mapped from Czech to the English words. Their result support the linguistic hypothesis. \newcite{belinkov-etal-2017-neural} train machine translation models between morphologicaly rich and morphologicaly poor languages, and test the model on morphological tasks. This is also a related effort. In addition, this question is closely related to the multilingual parsing task \cite{zeman-etal-2018-conll}. More work can be conducted in this direction, and multilingual parsing tasks can be a good reference.

\section{Conclusion}
\label{sec:conclude}

Neural network approaches have been applied to computational morphology, including both generation and analysis tasks, with great success for concatenative as well as non-concatenative morphology systems. The current state-of-the-art architecture is the Transformer. Before the Transformer, the most commonly used neural network architecture is (bidirectional) RNN with attention. The two gated RNN architectures -- LSTM and GRU -- have comparable performance though LSTM have been adopted in relatively more work. The feedforward neural network is also commonly used in computational morphology. The use of CNN is quite limited, used mainly to get better word embedding representations from characters. For morphological generation tasks, the encoder-decoder architecture has been dominant. Before the Transformer was successfully applied, the RNN-based encoder-decoder model with soft attention mechanism produced state-of-the-art performance when the training data size is large. Hard attention mechanism leverages the fact that the transduction in morphological generation has more copying than in machine translation. The seq2seq model with hard attention can tackle the generation tasks better when the training data is limited and achieve performance as good as or even better  than soft attention-enhanced seq2seq architecture in high-resource situations. The Transformer model has recently been applied to the morphological (re)inflection task, and produced the current state-of-the-art performance, even when the training data is limited \cite{wu2020applying,vylomova-etal-2020-sigmorphon,liu-hulden-2020-analogy}. 

The advantage of neural network approaches is not only much better performance with little high-level feature engineering. With these approaches, end-to-end learning becomes convenient, eliminating the error transmission between steps in the pipeline setting commonly used before deep learning methods. What's more, neural network approaches make multi-task and joint learning more doable. In addition, it becomes possible to model and explore research questions which are hard to study before, such as cross-lingual transfer and modeling of the non-linearity in human language acquisition, etc. Morphology in sentential context is also tackled more with better performance, thanks to neural network models which can represent context and meaning more effectively with dense vector representations.

Though data-hungriness has been a problem with neural network models, with recent efforts to improve neural models in low-resource settings and data augmentation techniques, neural models have achieved as good or even better performance when the training data is limited, compared to more traditional machine learning approaches. 

However, the neural network approach is not without problems. The lack of interpretability has made it hard to contribute to the understanding of human language, which is one of the focus of computational morphology. Current effort to interpret neural models has been trying to explain what these models have learned, but how humans understand and process language may be different from these models. It is questionable whether being able to speculate what neural models learn can shed light on human language understanding and generation. 

Another way neural models can contribute to the study of language, or morphology in the context of this paper, is to generate high-quality predictions even when the annotated data is limited, since the data in linguistic studies is usually limited compared to the large amount of data NLP tasks require. However, the performance of neural models in low-resource situations is not as good. In addition, the extreme situation of being low-resource is where there is no labeled data at all, a condition where we need to conduct unsupervised learning or reinforcement learning. There have been some unsupervised learning  work in computational morphology before neural models, but little work has been done for unsupervised learning or reinforcement learning of morphology with neural models. Future work is needed to solve these problems.

\label{LastPage}
\newpage

\pagenumbering{roman}
\pagestyle{fancy}
\fancyhf{} 
\renewcommand{\headrulewidth}{0pt} 
\cfoot{\thepage}

\bibliographystyle{acl}
\bibliography{bibliography}

\end{document}